\documentclass[acmtog,nonacm]{acmart}
\usepackage{subcaption}
\usepackage{enumitem}
\usepackage{cleveref}

\Crefname{figure}{Fig.}{Figs.}
\begin{document}
\title{UltraTex: Unleashing 2K Multi-View Diffusion for 3D Texturing}

\begin{teaserfigure}
  \includegraphics[width=\textwidth]{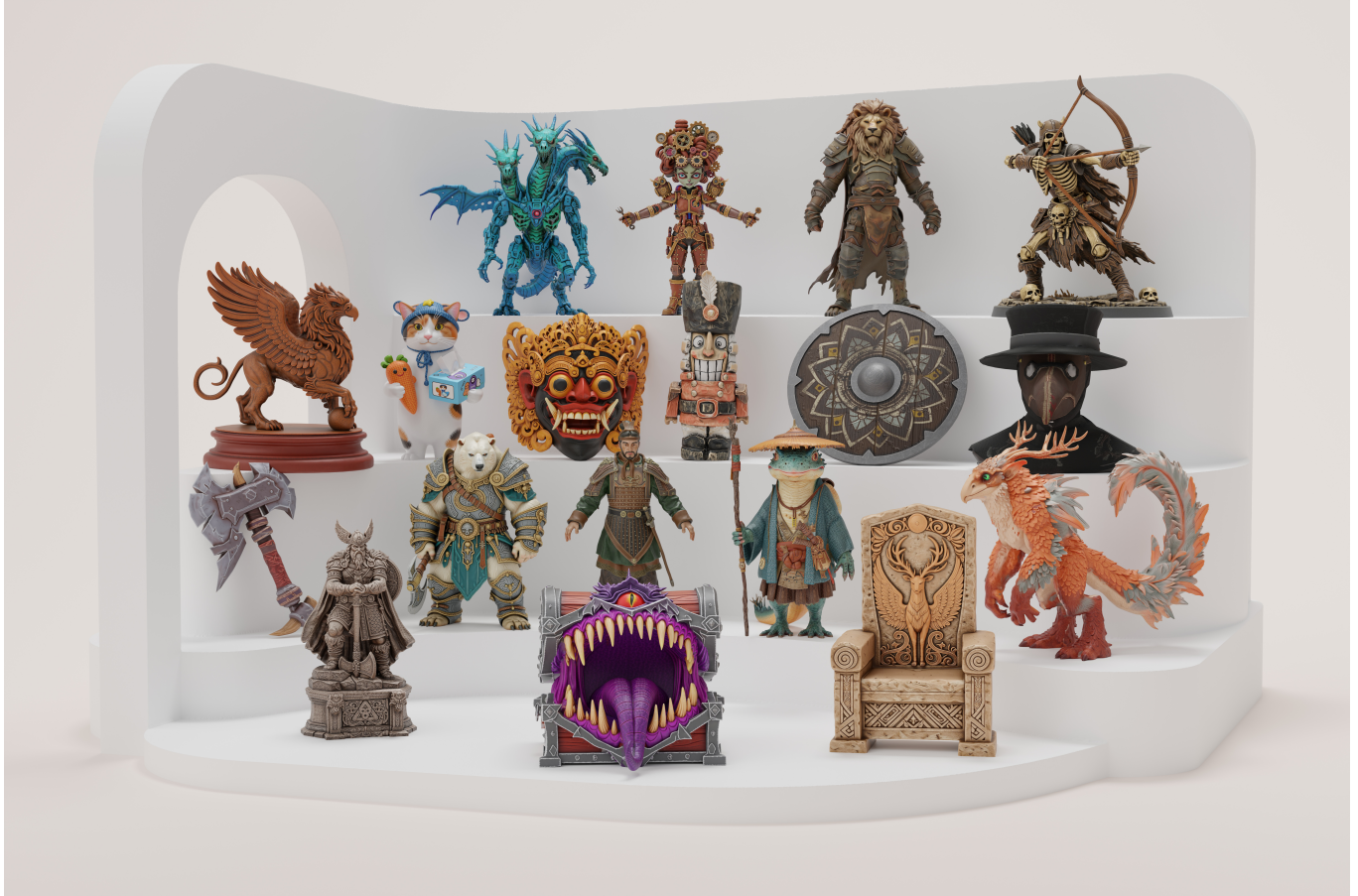}
  \caption{A gallery of diverse 3D assets textured by UltraTex, showcasing high-fidelity results with rich fine-grained details.}
  \label{fig:teaser}
\end{teaserfigure}

\author{Yibo Zhang}
\affiliation{%
  \institution{Jilin University}
  \city{Changchun}
  \country{China}}
\affiliation{%
  \institution{Shanghai Innovation Institute}
  \city{Shanghai}
  \country{China}}
\email{ybzhang23@mails.jlu.edu.cn}

\author{Ze Yuan}
\affiliation{%
  \institution{The University of Hong Kong}
  \city{Hong Kong}
  \country{China}}
\email{yuanze1024@connect.hku.hk}

\author{Nan Cao}
\affiliation{%
  \institution{Tongji University}
  \city{Shanghai}
  \country{China}}
\affiliation{%
  \institution{Shanghai Innovation Institute}
  \city{Shanghai}
  \country{China}}
\email{nan.cao@gmail.com}

\author{Li Zhang}
\affiliation{%
  \institution{Fudan University}
  \city{Shanghai}
  \country{China}}
\affiliation{%
  \institution{Shanghai Innovation Institute}
  \city{Shanghai}
  \country{China}}
\email{lizhangfd@fudan.edu.cn}

\author{Yan-Pei Cao}
\affiliation{%
  \institution{VAST}
  \city{Beijing}
  \country{China}}
\email{caoyanpei@gmail.com}

\author{Yuan-Chen Guo}
\affiliation{%
  \institution{VAST}
  \city{Beijing}
  \country{China}}
\email{imbennyguo@gmail.com}

\author{Rui Ma}
\authornote{Corresponding author.}
\affiliation{%
  \institution{Jilin University}
  \city{Changchun}
  \country{China}}
\email{ruim@jlu.edu.cn}

\keywords{3D texturing, multi-view image generation}

\begin{abstract}
High-quality texture generation is essential for creating realistic and production-ready 3D assets. Recent multi-view diffusion methods have shown promising results for image-guided 3D texturing, but they are typically constrained to low operating resolutions such as 512 or 768, making it difficult to preserve high-frequency details from high-resolution reference images. Scaling this paradigm to 2048 resolution is computationally prohibitive, as the unified multi-view sequence exceeds 212K tokens and incurs excessive memory and latency. In this paper, we present UltraTex, an efficient end-to-end framework for high-resolution multi-view diffusion-based 3D texturing. Our key observation is that object-centric multi-view renderings contain two major sources of redundancy: background-induced sequence redundancy and sparse token interactions within the foreground. To address them, we introduce Background Token Dropping, which removes background tokens before the DiT backbone, and Block-Sparse Attention, which reduces attention computation over the retained foreground sequence. 
To enable efficient foreground-only inference while avoiding reconstruction artifacts, we further design Foreground-Aware VAE Decoding to ensure the quality of the final high-resolution views.
To satisfy the demanding data requirements of 2K-resolution multi-view diffusion training, we construct G-buffer TexVerse, a large-scale, ultra-high-resolution multi-view rendering dataset covering over 268,000 3D assets.
Extensive experiments show that UltraTex generates visually faithful textures with rich fine-grained details, while substantially improving efficiency, achieving $20.6\times$--$91.1\times$ training speedup and $22.3\times$--$74.6\times$ end-to-end inference speedup over the baseline on common samples in our dataset. Code and data is at \color{blue}{\url{https://yiboz2001.github.io/UltraTex}}\color{black}{.}
\end{abstract}

\maketitle

\section{Introduction}

The automated generation of 
high-fidelity 3D assets is a central pillar of modern computer graphics, with widespread applications in gaming, film production, and spatial computing. While significant strides have been made in 3D geometry generation, producing visually compelling, production-ready textures remains a formidable challenge that is still heavily reliant on manual authoring.

Recently, multi-view diffusion models have emerged as the dominant paradigm for 3D texture generation
~\cite{MV-Adapter, Hunyuan3D-2.1, Hunyuan3D-2.5, Step1X-3D, Seed3D-1.0, AlignTex, UniTeX, CaliTex, LumiTex}.
By lifting strong visual priors from large-scale pretrained text-to-image or video diffusion models, these methods generate multi-view images that are subsequently reprojected onto 3D surfaces.
However, a glaring limitation persists: existing frameworks are fundamentally constrained to low operating resolutions 
(e.g., 512 or 768). 
Consequently, when provided with high-resolution reference images, these methods fail to preserve crucial high-frequency details, leading to blurred or globally inconsistent textures that fall short of production standards.


The primary roadblock to scaling this paradigm lies in the sheer computational complexity of high-resolution multi-view generation. Modern multi-view texturing~\cite{UniTeX, LumiTex, Seed3D-1.0} requires jointly modeling target noisy tokens, geometry-conditioning tokens, and reference-image tokens. Existing approaches typically concatenate these into a unified sequence and process them through a Diffusion Transformer (DiT)~\cite{DiT} using full self-attention. While viable at lower resolutions, this dense paradigm becomes computationally intractable as resolution increases. Specifically, targeting a canvas resolution of 2048 yields an input sequence exceeding 212,000 tokens. Processing such an extreme sequence length through deep feed-forward and attention layers incurs prohibitive GPU memory consumption and latency, making both training and inference practically impossible without a fundamental architectural rethinking.

Our core insight is that this computational bottleneck is heavily driven by two distinct forms of spatial redundancy inherent to object-centric multi-view rendering. First, we identify \textit{background-induced sequence redundancy}. In multi-view layouts, the foreground object typically occupies a highly variable and often small fraction of the canvas. Yet, standard DiTs process these vast expanses of empty, non-texture background space with the exact same computational budget as the texture-rich foreground, leading to catastrophic token-level waste. Second, we observe \textit{attention redundancy}. Even after eliminating the background, the remaining foreground tokens exhibit highly sparse attention patterns, demonstrating that computing dense, all-to-all attention across the compressed sequence remains largely unnecessary for local texture synthesis.

Building on these insights, we propose \textbf{{UltraTex}}, a highly efficient, end-to-end multi-view diffusion framework capable of generating 2048-resolution textures with uncompromising detail. To eliminate sequence redundancy, we introduce \textit{Background Token Dropping}. By leveraging geometric masks to discard background tokens before they enter the DiT backbone, we drastically shorten the sequence, ensuring that computation is exclusively dedicated to the foreground. To tackle attention redundancy, we apply \textit{Block-Sparse Attention} with a top-$k$ selection mechanism over this retained foreground sequence. Together, these complementary designs seamlessly bypass dense layer-wise modeling over the full canvas, concentrating computing power strictly on regions that dictate the final 3D texture.

While token dropping unlocks massive efficiency gains, it introduces a unique challenge during inference. To preserve speed at test time, we apply a foreground-only denoising strategy; however, this leaves the background in its initial Gaussian noise state. Directly feeding this composite latent, i.e., a clean denoised foreground juxtaposed with pure noise, into a standard VAE decoder produces severe artifacts and quality degradation in the reconstructed foreground. To solve this, we design \textit{Foreground-Aware VAE Decoding}. By substituting the noisy background with a canonical in-distribution background latent and lightly fine-tuning the VAE decoder with a foreground-constrained objective, we ensure robust, artifact-free reconstruction of the final high-resolution views.

Finally, training a 2K-resolution multi-view diffusion model requires data of unprecedented scale and quality. To support UltraTex, we construct G-buffer TexVerse, a rigorous, large-scale multi-view rendering dataset built upon TexVerse~\cite{TexVerse}. Covering over 268,000 meticulously filtered high-quality 3D assets, the dataset provides multi-view G-buffer attribute maps alongside reference views and shaded image sets rendered under diverse lighting conditions at up to \(4096 \times 4096\) resolution. This dataset provides the essential, structured data foundation required for high-resolution texture generation.

Extensive experiments demonstrate that UltraTex achieves state-of-the-art multi-view texture generation, producing detailed and visually faithful 3D assets while maintaining high computational efficiency. 
Meanwhile, on common samples in our dataset, our method achieves $20.6\times$--$91.1\times$ training speedup and $22.3\times$--$74.6\times$ end-to-end inference speedup over the baseline.
Our main contributions are summarized as follows:

\begin{enumerate}[itemsep=0.1em, topsep=0.2em, leftmargin=2em]
    \item We present UltraTex, an efficient framework 
    that successfully scales multi-view diffusion to 2048 resolution, unlocking the generation of high-fidelity, highly detailed 3D textures.
    \item We introduce a principled, foreground-aware computational design to eliminate the bottlenecks of high-resolution DiTs. This includes Background Token Dropping to compress sequence length, Block-Sparse Attention to optimize token interactions, and Foreground-Aware VAE Decoding to preserve inference efficiency without visual artifacts.
    \item We construct G-buffer TexVerse, a large-scale, ultra-high-resolution multi-view rendering dataset covering over 268,000 3D assets, providing structured and rigorously filtered training data for future 3D texturing research.
\end{enumerate}
\section{Related Work}

\subsection{3D Texturing via Multi-View Reprojection}
Recently, diffusion-model-based multi-view generation has become a dominant paradigm for 3D texture generation \cite{MV-Adapter, Hunyuan3D-2.1, Hunyuan3D-2.5, Step1X-3D, Seed3D-1.0, AlignTex, UniTeX, CaliTex, LumiTex}. 
Given a reference image and 3D geometry, these methods leverage the visual priors encoded in pretrained image or video diffusion models to generate multi-view texture images, which are then projected onto the surface for texturing. 
They have achieved significant progress in both generation quality and generalization capability. 
Some studies further explore cross-view information interaction to improve generation quality and view consistency \cite{AlignTex, CaliTex, MV-Adapter}.
However, existing methods typically support only relatively low-resolution inputs, making it difficult to fully preserve the fine details and high-frequency appearance information contained in high-resolution reference images. This limitation ultimately restricts the resolution, realism, and detail expressiveness of the generated textures. 
Meanwhile, efficient computational mechanisms for high-resolution multi-view diffusion remain insufficiently explored. 
To address this bottleneck, UltraTex introduces a foreground-aware efficient computation strategy that supports multi-view texture generation at 2048 resolution. While maintaining high computational efficiency, it enables high-quality, high-fidelity, and detail-rich texture generation for 3D assets.

\subsection{Efficient Multi-View Generation}
Existing efforts have largely focused on reducing the cost of cross-view attention in multi-view diffusion by redesigning the attention computation itself. Geometry-aware methods exploit camera or scene priors to restrict attention to more meaningful cross-view regions, for example by aligning epipolar lines with image rows under orthographic assumptions~\cite{ERA3D}, sampling sparsely along epipolar lines with Pl\"ucker ray embeddings~\cite{EpiDiff, SPAD}, or establishing pixel correspondences through a coarse proxy mesh~\cite{MEAT}.
More recently, CTR3D~\cite{CTR3D} compresses multi-view tokens into a smaller set of representative tokens within the attention layer and recovers them afterwards. These strategies effectively alleviate the burden of dense cross-view attention, but they mainly operate inside the attention module and leave the full token sequence unchanged throughout the DiT backbone.

In our task setting, the raw sequence length exceeds 200K tokens, incurring substantial memory and training-efficiency costs that go beyond what attention-side optimization alone can handle.
We therefore introduce Background Token Dropping, which leverages the geometric foreground mask to discard background tokens before they enter the DiT, fundamentally shortening the input fed into every DiT block. On top of this compressed foreground sequence, we further apply Block-Sparse Attention to reduce the residual attention cost. 
Together, the two designs act in a complementary manner, jointly reducing memory and compute throughout the DiT backbone.

\subsection{Sparse Attention}
The $O(N^2)$ complexity of full attention is a central bottleneck for long-sequence modeling. 
In long-context language modeling, this is mitigated through trainable sparse attention mechanisms~\cite{NSA, MoBA}. 
In visual diffusion generation, related work can be broadly grouped into two categories: \emph{training-free} methods leave model weights untouched and skip redundant computation at inference by analyzing or predicting sparse patterns in pretrained models~\cite{Sparse-vDiT, Radial-Attention, SpargeAttn}; \emph{trainable} methods jointly optimize the sparse mechanism with the model, spanning block-sparse attention, hierarchical selection, and other forms~\cite{VSA, VMoBA, LLSA, SLA}. For settings with more pronounced 3D or temporal structure, dedicated spatial or temporal sparse designs have also been proposed~\cite{Direct3D-S2, Sculpt4D}. 
In our method, Block-Sparse Attention operates on the compressed foreground sequence produced by Background Token Dropping, rather than on the original full-resolution sequence. Thus, it complements foreground token reduction by further reducing attention computation among the retained foreground tokens.

\subsection{Token Reduction}
Token reduction has been widely explored in Vision Transformers and Diffusion Transformers to reduce computational cost by removing less informative tokens~\cite{DynamicViT,EViT,LVTP, DyDiT}. Most existing methods determine token importance using learned predictors, attention scores, or intermediate features, and perform pruning dynamically within Transformer layers. In contrast, our method exploits explicit geometric masks to deterministically identify and remove invalid tokens without learning additional token-importance modules. Moreover, token dropping is performed once before the MM-DiT backbone, such that all subsequent Transformer layers directly operate on the reduced token sequence. This design is particularly suited to multi-view texture generation, where geometric masks provide explicit prior knowledge of irrelevant background regions.

\begin{figure*}[t]
    \centering
    \includegraphics[width=0.98\textwidth]{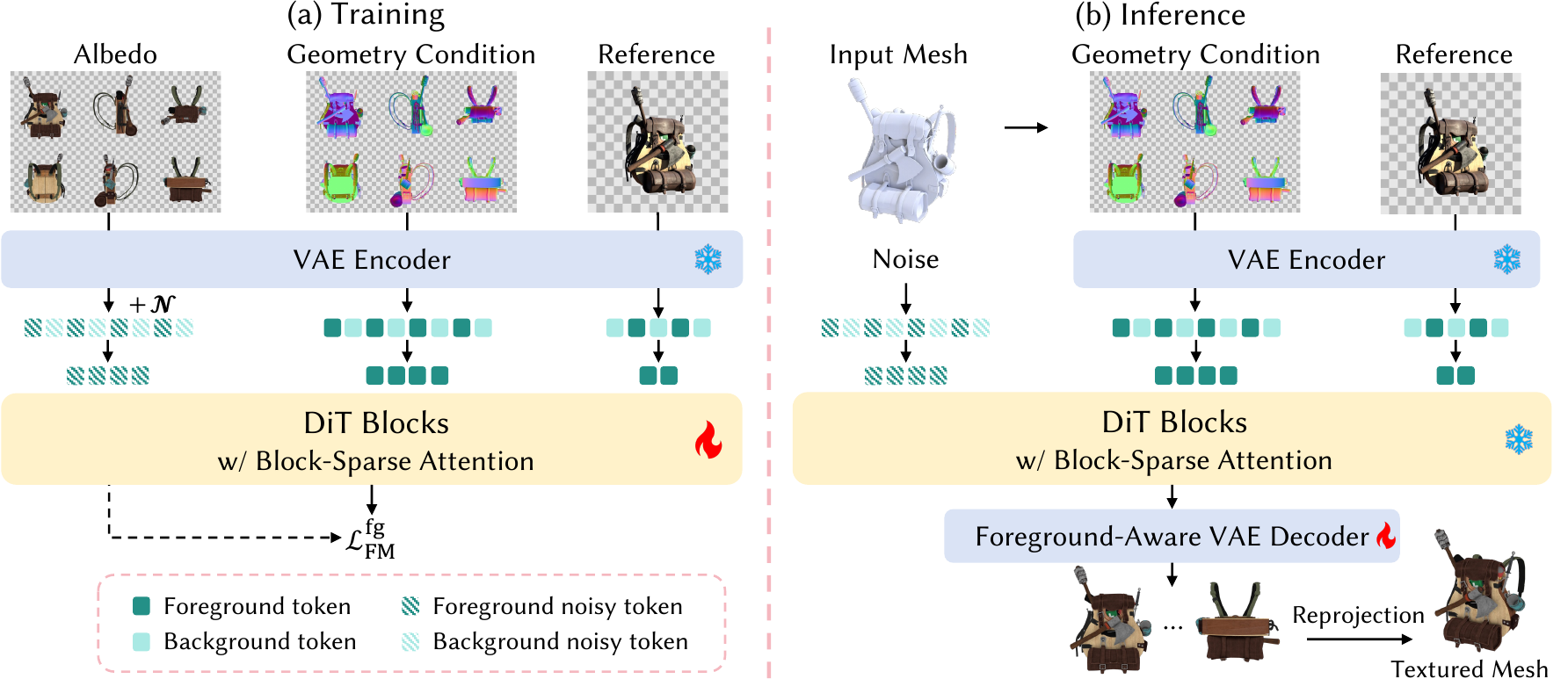}
    \caption{
        Overview of UltraTex.  
        UltraTex generates geometry-aligned 2048-resolution multi-view texture images for 3D texturing.
        UltraTex first removes all background tokens before the DiT blocks, and then applies block-sparse attention on the remaining foreground sequence. During inference, foreground-only denoising and foreground-aware VAE decoding generate multi-view results, which are reprojected onto the input mesh to obtain the final textured 3D asset.
    }
    \label{fig:pipeline}
\end{figure*}
\section{Methodology}
\label{sec:method}

Given a reference image $\mathbf{I}_\text{ref}$ and six per-view normal maps $\mathbf{I}_\text{cond}$ as geometric guidance at an ultra-high resolution of $2048\times2048$, UltraTex aims to generate geometry-aligned, multi-view texture images through an end-to-end diffusion framework.
These generated views serve as high-resolution texture observations that are subsequently reprojected to create the final textured 3D asset. 
\Cref{fig:pipeline} illustrates the overall pipeline.
To overcome the prohibitive computational barriers of modeling 2K-resolution distributions, we decompose the generative bottleneck into three distinct levels of spatial and computational redundancy, addressing each systematically:
\begin{enumerate}[itemsep=0.1em, leftmargin=2em]
    \item \textbf{Sequence-Level}: We introduce \textit{Background Token Dropping} (Sec.~\ref{sec:token_dropping}) to eliminate non-texture regions, drastically compressing the sequence length before the transformer backbone.
    \item \textbf{Attention-Level}: Over the compressed sequence, we apply \textit{Block-Sparse Attention} (Sec.~\ref{sec:sparse_attention}) to bypass dense, all-to-all interactions among the remaining foreground tokens.
    \item \textbf{Decoding-Level}: To maintain boundary fidelity during fast, foreground-only inference, we introduce \textit{Foreground-Aware VAE Decoding} (Sec.~\ref{sec:vae_decoding}).
\end{enumerate}

\subsection{Base Architecture: In-Context Multi-View Diffusion}
\label{sec:overview}

Our generator is built upon the pretrained FLUX  model~\cite{FLUX}, which adopts the Multi-Modal Diffusion Transformer (MM-DiT) architecture~\cite{MMDiT} and follows the Flow Matching framework~\cite{Flow-Matching}.
To cast 3D texturing as a conditional generative task, we utilize an in-context conditioning paradigm. Specifically, the reference image $\mathbf{I}_{\text{ref}}$, multi-view geometric normals $\mathbf{I}_{\text{cond}}$, and ground-truth multi-view albedo images $\mathbf{I}_{\text{tgt}}$ are first encoded into latent token sequences $\mathbf{z}_{\text{ref}}$, $\mathbf{z}_{\text{cond}}$, and $\mathbf{z}_{\text{tgt}}$, respectively. The model operates on a unified sequence constructed by concatenating these tokens.
Following the Flow Matching formulation, for a given timestep $t \in [0, 1]$ and Gaussian noise $\epsilon \sim \mathcal{N}(0, \mathbf{I})$, we construct the noisy target tokens as:
\begin{equation}
z_\text{tgt}^{(t)} = (1-t) z_\text{tgt} + t \varepsilon .
\label{eq:z_tgt_t}
\end{equation}
Under this data-to-noise path, the target velocity is $\varepsilon - z_\text{tgt}$, and only the target tokens are supervised for velocity prediction. The model is trained with the standard flow matching objective:
\begin{equation}
    \mathcal{L}_\text{FM}(\theta)
    =
    \mathbb{E}_{t,\,z_\text{tgt},\,\varepsilon}
    \left[
    \left\|
    \hat{v}_\theta(z_\text{tgt}^{(t)}, z_\text{ref}, z_\text{cond}, t)
    -
    (\varepsilon - z_\text{tgt})
    \right\|_2^2
    \right].
    \label{eq:flow_matching}
\end{equation}

While this full-attention DiT architecture is highly effective at lower resolutions, scaling it exposes a fundamental flaw. Operating at a $2048 \times 2048$ resolution results in $16,384$ tokens per image (after $8\times$ VAE downsampling and $2 \times 2$ DiT patchification). A standard six-view generation setup yields a staggering combined sequence length: $6 \times 16,384 \text{ (noisy latents)} + 6 \times 16,384 \text{ (conditions)} + 16,384 \text{ (reference)} = \mathbf{212,992 \text{ tokens}}$. Processing this sequence through deep transformer blocks incurs catastrophic GPU memory usage and intractable training times, motivating our foreground-aware architectural redesign.

\subsection{Background Token Dropping: Eliminating Sequence Redundancy}
\label{sec:token_dropping}

\begin{figure}[t]
    \centering
    \includegraphics[width=0.475\textwidth]{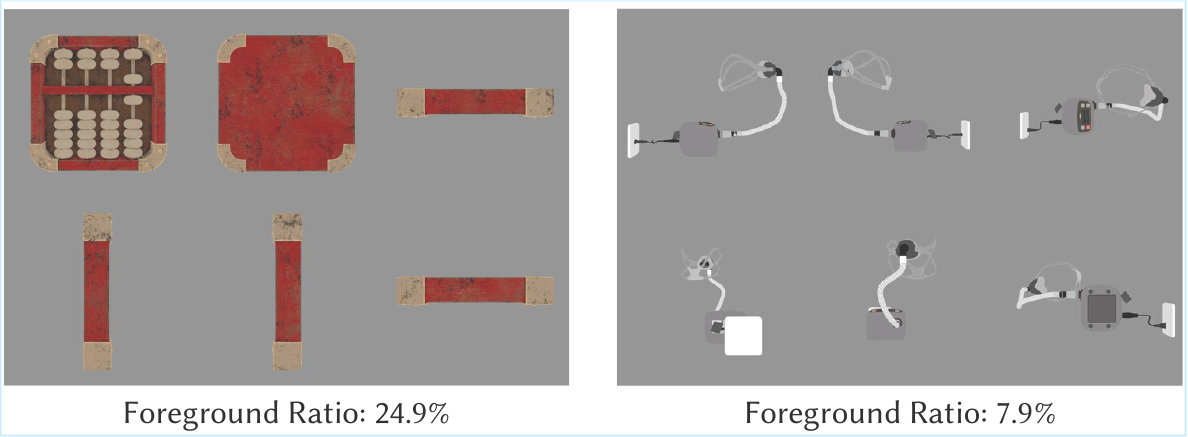}
    \vspace{-0.5cm}
    \caption{Examples of foreground sparsity in object-centric multi-view renderings. Each image shows a six-view layout, where only the object foreground corresponds to surface regions requiring texture generation.
    The two examples contain only $24.9\%$ and $7.9\%$ foreground pixels, respectively, highlighting the substantial background redundancy.
    }
    \label{fig:foreground_ratio}
\end{figure}
\begin{figure}[t]
    \centering
    \includegraphics[width=0.475\textwidth]{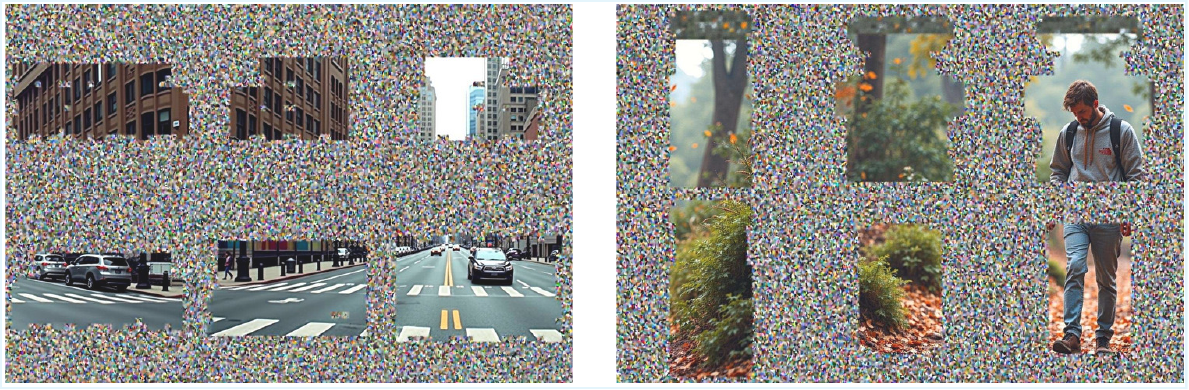}
    \vspace{-0.5cm}
    \caption{
        Preliminary token-dropping experiment with pretrained FLUX.
        We remove background tokens according to the foreground masks in the multi-view layout while preserving the original positional indices of the retained tokens.
        The model can still denoise the resulting discontinuous partial sequence and produce spatially plausible foreground results.
    }
    \label{fig:FLUX_token_drop}
\end{figure}

A key property of object-centric multi-view rendering is that the target geometry occupies vastly different spatial extents across different viewpoints. As illustrated in~\Cref{fig:foreground_ratio}, a substantial majority of the canvas often corresponds to empty background space. For example, specific views may contain only 7.9\% to 24.9\% valid foreground pixels. In standard DiT pipelines, these background regions are indiscriminately converted into tokens and processed by every attention and feed-forward layer, leading to profound computational waste.

\paragraph{Spatial Consistency via Positional Embeddings.} This observation raises a fundamental question: \textit{Can we entirely discard background tokens without destroying the model's structural understanding of the 2D canvas?} We conducted an exploratory experiment using the pretrained FLUX inference pipeline. We provide a multi-view layout foreground mask, injected positional embeddings into the initial noise, and subsequently deleted all background tokens. Remarkably, as shown in \Cref{fig:FLUX_token_drop}, performing denoising solely on this discontinuous, partial token sequence still yields spatially plausible results. This proves that the DiT relies on its Rotary Position Embeddings (RoPE)~\cite{RoPE}, not the contiguity of the 1D token sequence, to understand spatial layout.

Guided by this insight, we propose \textit{Background Token Dropping}. By stripping away background tokens \textit{before} they enter the MM-DiT, we force the network to allocate 100\% of its computational budget to regions that actively define the 3D surface texture.

\paragraph{Mask Construction and Sequence Compression.}
We extract binary foreground masks directly from the alpha channels  of the rendered multi-view data. These masks are downsampled to $1/16$ resolution to align with the latent token grid and slightly dilated to preserve object boundary details.
Let $\mathcal{M}_{\text{fg}}$ denote the set of retained foreground positions. 
The six per-view masks are applied to both the noisy target and the geometric conditioning tokens, while the reference image utilizes its own distinct  foreground mask.
After dropping the background, the retained foreground tokens are concatenated to form the compressed MM-DiT input.
Crucially, each retained token carries its original RoPE index. 
This preserves the absolute spatial coordinates of every pixel, seamlessly maintaining the strict epipolar and geometric priors required for multi-view consistency.

\paragraph{Foreground-Restricted Training.}
During training, we construct the noisy target tokens $z_\text{tgt}^{(t)}$ on the full dense latent grid as in Eq.~\ref{eq:z_tgt_t}, and subsequently apply token-dropping mask. Let
$
\tilde{z}_\text{tgt}^{(t)}
=
\mathrm{Gather}_{\mathcal{M}_\text{fg}}\!\left(z_\text{tgt}^{(t)}\right)
$
denote the compacted noisy sequence. The model takes $\tilde{z}_\text{tgt}^{(t)}$ together with the similarly masked condition and reference tokens,  and predicts velocities \textit{only} for these foreground target tokens.
The flow matching objective is inherently reformulated to ignore empty space:
\begin{equation}
\mathcal{L}_\text{FM}^{\text{fg}}(\theta)
=
\mathbb{E}_{t,\, z_\text{tgt},\, \varepsilon}
\left[
\left\|
\hat{v}_\theta^\text{fg}
-
\mathrm{Gather}_{\mathcal{M}_\text{fg}}
\left(\varepsilon - z_\text{tgt}\right)
\right\|_2^2
\right],
\label{eq:loss_fm}
\end{equation}
where $\hat{v}_\theta^\text{fg}$ denotes the model prediction on $\tilde{z}_\text{tgt}^{(t)}$.

\paragraph{Efficient Inference.}
At test time, we apply the same foreground-only denoising strategy to fully realize the massive latency reductions.
Starting from Gaussian noise, UltraTex selectively predicts velocities exclusively for the positions defined by $\mathcal{M}_{\text{fg}}$, and the Euler update is applied sparsely:
\begin{equation}
z^{(t_\text{prev})}_{i}
=
z^{(t_\text{curr})}_{i}
+
(t_\text{prev}-t_\text{curr})\,\hat{v}_{\theta,i},
\quad \forall i \in \mathcal{M}_\text{fg}.
\label{eq:euler}
\end{equation}
Background positions are not updated and therefore remain in their initial Gaussian noise state throughout the denoising process.

\begin{figure}[t]
    \centering
    \includegraphics[width=0.475\textwidth]{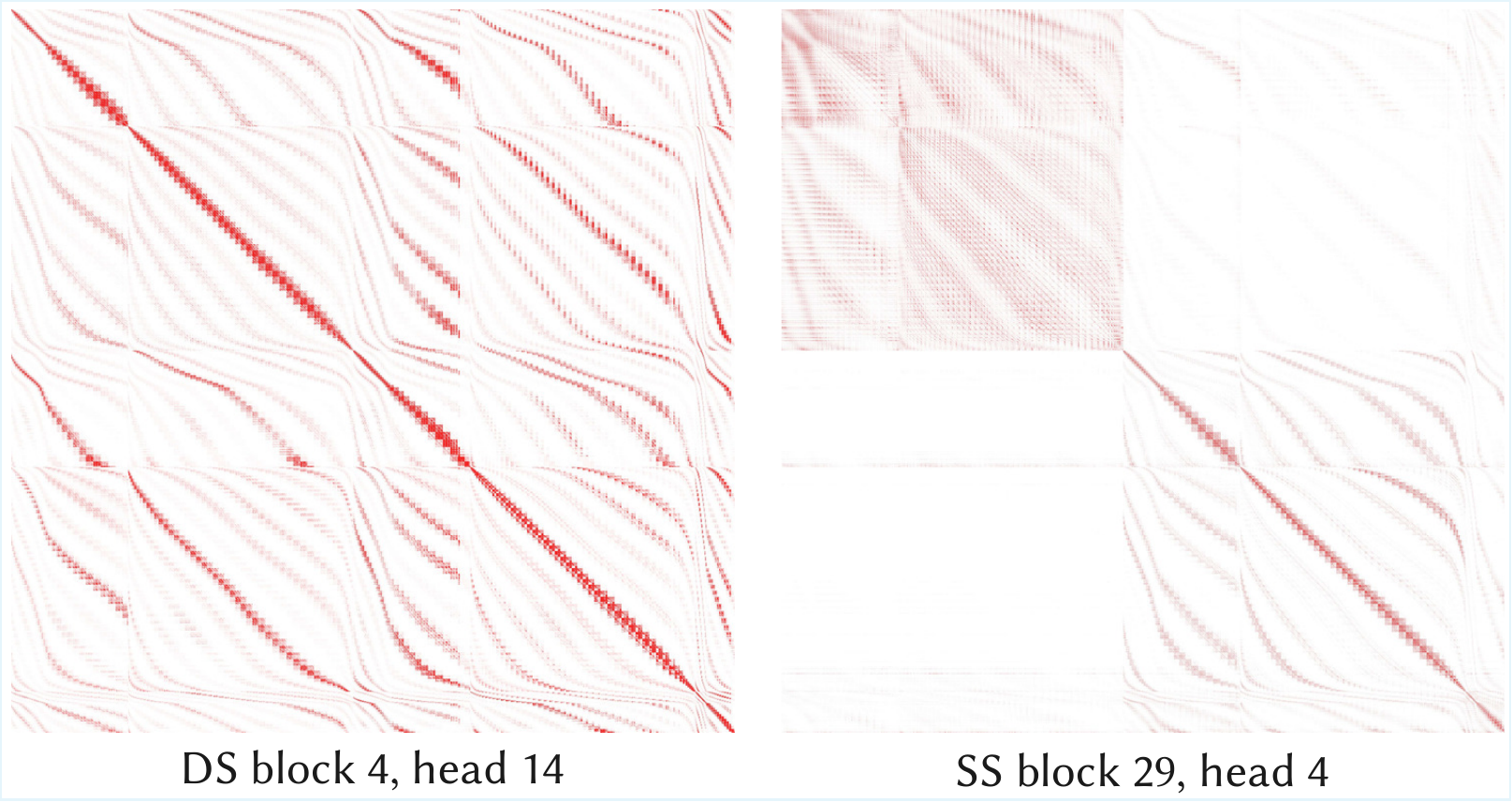}
    \caption{
        Attention sparsity over the retained foreground sequence after Background Token Dropping.
        We visualize attention maps from a lower-resolution full-attention model trained with Background Token Dropping.
        Both Double-Stream (DS) and Single-Stream (SS) DiT block attention maps exhibit sparse patterns over the retained foreground tokens.
    }
    \label{fig:attention_drop}
\end{figure}
\subsection{Block-Sparse Attention: Mitigating Attention Redundancy}
\label{sec:sparse_attention}
After dropping background token, the dominant redundancy from background regions is removed and the input sequence is substantially shortened.
Nevertheless, at $2048 \times 2048$ resolution, dense attention among the retained foreground tokens still accounts for a considerable portion of the training and inference cost.
We therefore further examine the attention computation over the retained foreground tokens.
Specifically, we train a full-attention multi-view generation model at lower resolutions with Background Token Dropping and analyze its attention maps.
As shown in \Cref{fig:attention_drop}, both double-stream blocks and single-stream blocks exhibit sparse attention patterns over the retained foreground tokens, suggesting that the remaining computation can be reduced by exploiting sparsity in the attention pattern itself.
Based on this observation, we further apply Block-Sparse Attention with a Top-K selection mechanism over the compressed foreground sequence.

Let \( Q, K, V \in \mathbb{R}^{N \times d} \) denote the query, key, and value matrices, where \(N\) is the sequence length after Background Token Dropping and \(d\) is the feature dimension of each attention head.
Full self-attention incurs a computational complexity of \( \mathcal{O}(N^2 d) \) due to two matrix multiplications:
\begin{equation}
S = \frac{QK^\top}{\sqrt{d}}, \quad
P = \mathrm{Softmax}(S) \in \mathbb{R}^{N \times N}, \quad
O = PV \in \mathbb{R}^{N \times d}.
\end{equation}
The tensors are first partitioned into query and key-value blocks:
\(
Q = \{\mathbf{Q}_i\}_{i=1}^{N_q}, \quad
K = \{\mathbf{K}_j\}_{j=1}^{N_k}, \quad
\)
where \( \mathbf{Q}_i \in \mathbb{R}^{b_q \times d} \), 
\( \mathbf{K}_j, \mathbf{V}_j \in \mathbb{R}^{b_{kv} \times d} \),
\(N_q = N / b_q\), and \(N_k = N / b_{kv}\).
To estimate block-level relevance with low overhead, we compute a compressed score matrix
\(S_c \in \mathbb{R}^{N_q \times N_k}\) from mean-pooled query and key blocks:
\begin{equation}
S_c = \mathrm{pool}(Q) \cdot \mathrm{pool}(K)^\top,
\end{equation}
where \( \mathrm{pool}(\cdot) \) denotes mean pooling within each block along the token dimension.
For each attention head and each query block \(i\), we perform row-wise Top-K selection over the key blocks.
Given a retention ratio \(\rho\), we retain
\(K_\rho = \lfloor \rho N_k \rfloor\) key blocks with the largest scores in \(S_c[i,:]\), producing a binary block mask \(M_c \in \mathbb{R}^{N_q \times N_k}\):
\begin{equation}
M_c[i,j] =
\begin{cases}
1, & j \in \operatorname{TopK}(S_c[i,:], K_\rho), \\
0, & \text{otherwise}.
\end{cases}
\end{equation}
Here, \(\operatorname{TopK}(S_c[i,:], K_\rho)\) returns the indices of the \(K_\rho\) key blocks with the largest scores for query block \(i\).
The compressed scores are used only to determine the sparse block layout.
Given the selected key-block set
\(\mathcal{S}_i = \{j \mid M_c[i,j]=1\}\), attention for query block \(i\) is computed using the original, unpooled \(Q\), \(K\), and \(V\):
\begin{equation}
\mathbf{O}_i =
\mathrm{Softmax}
\left(
\frac{\mathbf{Q}_i \mathbf{K}_{\mathcal{S}_i}^{\top}}{\sqrt{d}}
\right)
\mathbf{V}_{\mathcal{S}_i}.
\end{equation}
Thus, unselected query-key block pairs are skipped by the sparse attention kernel, reducing attention computation while preserving full token-level attention within the selected blocks.

\subsection{Foreground-Aware VAE Decoding: Preserving Inference Fidelity}
\label{sec:vae_decoding}

During inference, we apply foreground-only denoising to preserve the efficiency gains of \textit{Background Token Dropping}: only foreground latent positions are updated during sampling.
However, since background positions are not denoised, they remain in their initial Gaussian noise state throughout the sampling process.
Directly decoding the resulting composite latent, which consists of denoised foreground and noisy background, leads to noticeable quality degradation in the reconstructed foreground.
To address this issue, we design \textit{Foreground-Aware VAE Decoding}.
It consists of two steps: (1) replacing the noisy background latent with a canonical in-distribution background latent, and (2) lightly fine-tuning the decoder with a foreground-restricted objective to improve reconstruction fidelity on the resulting composite latent.

\paragraph{Canonical background latent.}
We first replace the noisy background with a canonical background latent obtained from a valid VAE encoding.
Specifically, we encode a solid-color black RGB image using the frozen VAE encoder to obtain $z^\text{bg}$.
This provides a valid, in-distribution VAE latent for the background region.
The latent is tile-repeated to cover the full $2\times3$ grid, and the composite latent is assembled as:
\begin{equation}
  z^\text{mix}_i = \begin{cases}
    z^\text{fg}_{0,i} & \text{if } i \in \mathcal{M}_\text{fg}, \\
    z^\text{bg}_i & \text{otherwise},
  \end{cases}
  \label{eq:mixed_latent}
\end{equation}
where $z^\text{fg}_{0}$ is the fully denoised foreground latent at timestep $t{=}0$ and subscript $i$ indexes the spatial position.
The canonical background latent replaces the noisy background with a more suitable decoder input, but this step alone may still leave foreground color shifts. We therefore further fine-tune the decoder to better adapt it to the composite latent $z^\text{mix}$.
 
\paragraph{Foreground-restricted decoder fine-tuning.}
We freeze the FLUX VAE encoder and perform lightweight fine-tuning on only the decoder, using a reconstruction loss restricted to foreground pixels:
\begin{equation}
  \mathcal{L}_\text{dec} = \frac{1}{|\mathcal{M}^\text{px}_\text{fg}|} \sum_{i \in \mathcal{M}^\text{px}_\text{fg}} \left\| \mathcal{D}(z^\text{mix})_i - x^\text{gt}_i \right\|^2,
  \label{eq:loss_dec}
\end{equation}
where $\mathcal{D}$ is the decoder, $x^\text{gt}$ is the ground-truth image, and $\mathcal{M}^\text{px}_\text{fg}$ is the foreground mask at full pixel resolution.
By applying supervision only to foreground pixels, the decoder is encouraged to focus on reconstructing high-quality foreground content while ignoring background regions in the loss.
This adaptation mitigates foreground color shifts and improves reconstruction fidelity.


\subsection{G-buffer TexVerse Dataset}
\label{sec:dataset}
To support high-resolution 3D texture generation, we construct G-buffer TexVerse, a large-scale multi-view rendering dataset built upon TexVerse~\cite{TexVerse}. It contains over 268K filtered 3D assets with multi-view G-buffer, reference images, and shaded observations rendered under diverse illumination conditions at resolutions up to $4096 \times 4096$. During training, we use only normal maps as geometric conditions and uniformly resize all rendered views to $2048 \times 2048$.
We further analyze the foreground-ratio distribution of the rendered dataset. More than $85\%$ of the assets fall within the $5\%$--$30\%$ foreground-ratio range, showing that object-centric multi-view renderings contain substantial background redundancy. Detailed asset curation, rendering protocol, data organization, and dataset statistics are provided in the supplementary material.

\section{Experiments}
\subsection{Implementation Details}
\paragraph{Training.}
We initialize the MM-DiT backbone from FLUX \cite{FLUX} and adapt it via LoRA with rank 64. Training proceeds on 64 H200 GPUs in three progressive stages, with a per-GPU batch size of 1.
Stage~1 trains at $512\times512$ with full attention and learning rate $1 \times 10^{-4}$ for 22K steps, taking 1 day.
Stage~2 increases the resolution to $1024\times1024$ and continues with full attention at learning rate $1 \times 10^{-5}$ from step 22K to 70K, taking 8 days.
Stage 3 further scales to $2048 \times 2048$ and switches to Block-Sparse Attention with the retention ratio set to $\rho=0.2$, using a learning rate of $2 \times 10^{-5}$ from step 70K to 97K, taking another 9 days.
To improve training efficiency, training samples are bucketed by their effective foreground token count, ensuring that GPUs within a data-parallel batch process sequences of similar length and reducing workload imbalance. 
To reduce memory overhead, we further adopt DeepSpeed ZeRO-2 \cite{DeepSeed} and gradient checkpoint \cite{GradientCheckpoint} for scalable training of the high-resolution multi-view diffusion model.

\paragraph{Foreground-aware VAE decoder fine-tuning.}
For Foreground-Aware VAE Decoding, we use the albedo data of training-set objects as supervision and uniformly resize it to a resolution of $2048 \times 2048$. Specifically, we fine-tune the FLUX VAE decoder for $3{,}000$ steps with a batch size of $64$ and a learning rate of $1 \times 10^{-5}$.

\begin{table*}[t]
\centering

\begingroup

\resizebox{\textwidth}{!}{
\begin{tabular}{l|ccccc|ccccc|ccccc}
\toprule
& \multicolumn{5}{c|}{\textbf{Unshaded}}
& \multicolumn{5}{c|}{\textbf{Shaded}}
& \multicolumn{5}{c}{\textbf{Relighting}} \\
\cmidrule(lr){2-6}
\cmidrule(lr){7-11}
\cmidrule(lr){12-16}

\textbf{Method}
& FID$\downarrow$ & CLIP-FID$\downarrow$ & CMMD$\downarrow$ & CLIP-I$\uparrow$ & LPIPS$\downarrow$
& FID$\downarrow$ & CLIP-FID$\downarrow$ & CMMD$\downarrow$ & CLIP-I$\uparrow$ & LPIPS$\downarrow$
& FID$\downarrow$ & CLIP-FID$\downarrow$ & CMMD$\downarrow$ & CLIP-I$\uparrow$ & LPIPS$\downarrow$ \\
\midrule

Step1X-3D
& -- & -- & -- & -- & --
& 160.3 & 19.265 & 0.613 & 0.900 & 0.154
& 160.7 & 19.189 & 0.600 & 0.901 & 0.154 \\

UniTEX
& -- & -- & -- & -- & --
& 133.8 & 16.750 & 0.524 & 0.918 & 0.116
& 133.4 & 15.697 & 0.458 & 0.923 & 0.113 \\

Hunyuan3D 2.1
& 153.48 & 18.995 & 1.150 & 0.908 & 0.136
& 123.1 & 14.354 & 0.432 & 0.929 & 0.119
& 126.0 & 14.117 & 0.409 & 0.930 & 0.117 \\

LumiTex
& 147.31 & 18.251 & 1.223 & 0.914 & 0.138
& 110.6 & 12.265 & \textbf{0.351} & 0.939 & 0.107
& 114.5 & 11.982 & \textbf{0.316} & 0.941 & 0.105 \\

UltraTex (Ours)
& \textbf{125.13} & \textbf{15.180} & \textbf{0.810} & \textbf{0.929} & \textbf{0.082}
& \textbf{97.6} & \textbf{12.239} & 0.368 & \textbf{0.941} & \textbf{0.064}
& \textbf{96.4} & \textbf{11.583} & 0.333 & \textbf{0.944} & \textbf{0.062} \\

\bottomrule
\end{tabular}
}

\caption{Quantitative comparison with baseline methods.
$\downarrow$ indicates lower is better, and $\uparrow$ indicates higher is better.}
\label{tab:quantitative_comparison2}

\endgroup
\end{table*}

\subsection{Comparisons}
\paragraph{Baselines}
We compare our method with several open-source state-of-the-art 3D texturing approaches, including Step1X-3D \cite{Step1X-3D}, Hunyuan3D 2.1 \cite{Hunyuan3D-2.1}, LumiTex \cite{LumiTex}, and UniTEX \cite{UniTeX}. 
Among them, Step1X-3D and UniTEX are texture-only methods whose generated textures contain baked-in lighting effects, while Hunyuan3D 2.1 and LumiTex are PBR-based methods that predict physically based material properties in addition to texture appearance. 

\paragraph{Metrics.}
Following previous works \cite{LumiTex,CaliTex,Hunyuan3D-2.1}, we adopt FID \cite{FID}, CLIP-FID \cite{CLIP}, CLIP-I \cite{CLIP}, and LPIPS \cite{LPIPS} to evaluate the visual fidelity and perceptual quality of the generated textures. In addition, we report CLIP Maximum Mean Discrepancy (CMMD) \cite{CMMD} to assess the distributional diversity of generated texture details.

\begin{table}[t]
\centering
\caption{Quantitative comparison of different VAE reconstruction settings.}
\label{tab:vae_comparison_table}
\begin{tabular}{lccc}
\toprule
Method & PSNR $\uparrow$ & SSIM $\uparrow$ & LPIPS $\downarrow$ \\
\midrule
FLUX VAE (Full Image)        & \underline{47.65} & \underline{0.9937} & \textbf{0.0023} \\
\midrule
FLUX VAE (Noise Background)  & 29.66 & 0.9555 & 0.0263 \\
Ours (w/o finetune)           & 37.81 & 0.9818 & 0.0084 \\
Ours (w/ finetune)            & \textbf{50.26} & \textbf{0.9956} & \underline{0.0038} \\
\bottomrule
\end{tabular}
\end{table}

\subsection{Quantitative Results}
We evaluate UltraTex on a held-out TexVerse test set comprising 100 objects unseen during training. 
We reconstruct the evaluation into three tracks: \textbf{Unshaded}, \textbf{Shaded}, and \textbf{Relighting}. The unshaded track evaluates methods that explicitly produce lighting-free albedo textures; for texture-only baselines with baked lighting, we do not report albedo metrics to avoid unfair comparison. For the shaded and relighting tracks, we render every method’s final textured mesh under the unified rendering pipeline. The shaded setting uses the same lighting as the input reference image, while the relighting setting uses randomly sampled environment maps. For baselines without PBR channels and our method, we use a fixed material setting with default roughness/metallic of 1/0. All tracks compare against ground truth over 32 viewpoints.
The corresponding results are shown in \Cref{tab:quantitative_comparison2}. 
UltraTex achieves the best performance on most metrics across the three tracks with fixed default material parameters for rendering, verifying that our lighting-free textures remain faithful under consistent rendering and novel lighting conditions.



\subsection{Qualitative Results}
\begin{figure*}[h]
    \centering
    \includegraphics[width=0.98\textwidth]{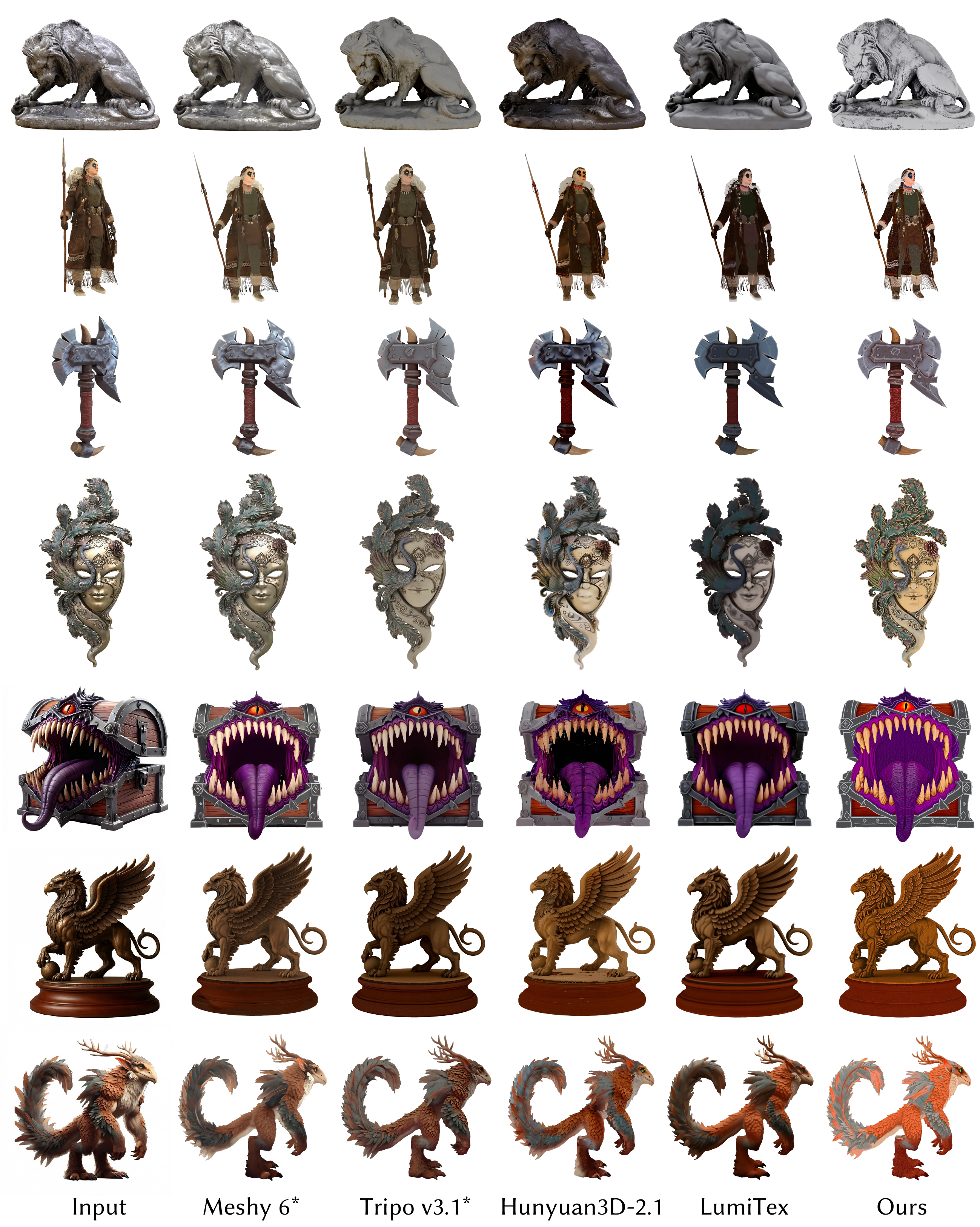}
    \caption{Qualitative comparison with state-of-the-art PBR-based methods (* denotes a commercial model). Our method consistently recovers finer texture details and achieves high visual quality across diverse inputs, including artist-created assets (first four examples) and in-the-wild images (last three examples).
    }
    \label{fig:qualitative_results}  
\end{figure*}

We evaluate UltraTex on diverse inputs, including artist-created 3D assets and in-the-wild images. For the in-the-wild cases, we obtain the input geometry using Tripo v3.1 \cite{tripo2026h31} and then apply different methods for image-guided texturing. 
For albedo generation, we compare UltraTex with state-of-the-art PBR-based methods \cite{Hunyuan3D-2.1,LumiTex}. 
As shown in \Cref{fig:qualitative_results}, existing methods often generate over-smoothed albedo textures and fail to recover fine local details. 
In contrast, UltraTex consistently preserving richer high-frequency details based on high-resolution geometric conditions while achieving better visual coherence.
We further include results from Meshy 6 \cite{meshy} and Tripo v3.1 commercial models as additional comparisons.

\subsection{Effect of Foreground-Aware VAE Decoding}
\Cref{fig:vae_recon} qualitatively compares different decoding strategies under foreground-only denoising.
As a standard VAE reconstruction reference, we reconstruct the complete ground-truth image using the original FLUX VAE.
In the foreground-only setting, directly decoding a composite latent consisting of a denoised foreground and a noisy background leads to noticeable foreground quality degradation, suggesting that noisy background latents are unsuitable inputs for the VAE decoder.
Replacing the noisy background with a latent encoded from a solid-color image provides a valid in-distribution background representation and substantially improves foreground reconstruction.
Furthermore, foreground-restricted fine-tuning better adapts the decoder to this composite latent structure and improves its robustness to foreground color shifts.
We further evaluate reconstruction performance on the object albedo data from our benchmark.
As shown in \Cref{tab:vae_comparison_table}, noisy background latents significantly degrade reconstruction quality, while our \textit{Foreground-Aware VAE Decoding} further improves reconstruction stability and fidelity under foreground-only denoising.

\subsection{Efficiency Analysis}
\begin{figure*}[!t]
      \centering
      \includegraphics[width=\linewidth]{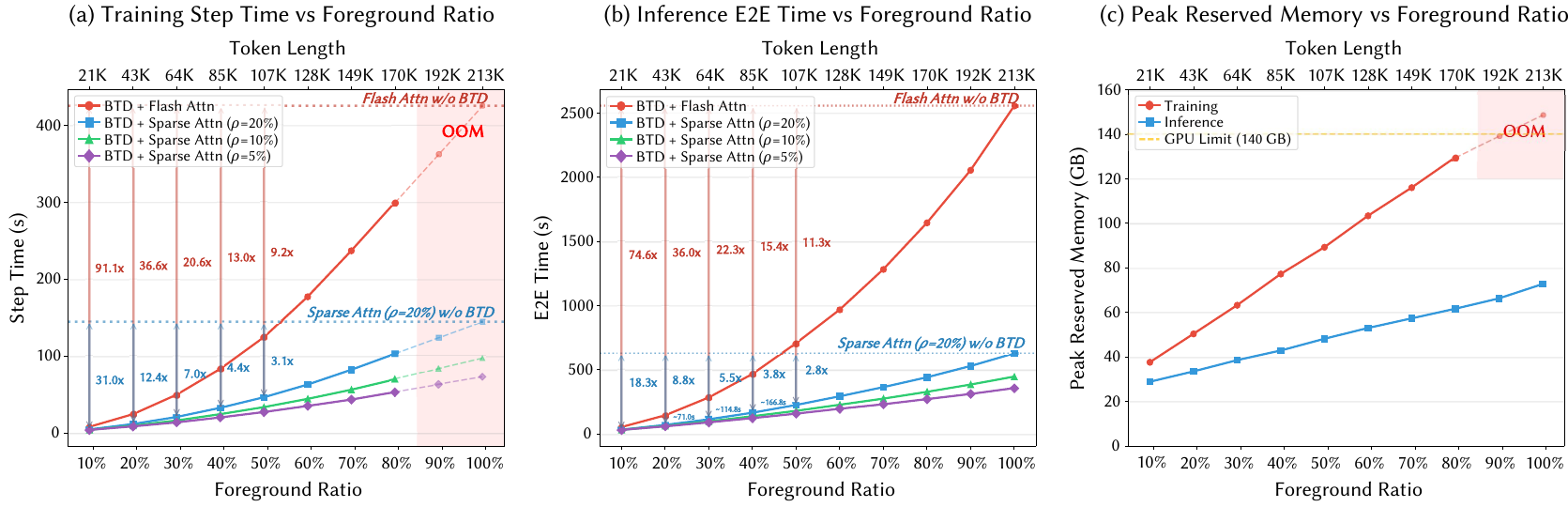}
      \caption{Efficiency analysis of UltraTex at 2048 resolution.
      Background Token Dropping (BTD) reduces training/inference cost and peak GPU memory by dropping background tokens and shortening the effective sequence,
   while Block-Sparse Attention further improves efficiency over the retained foreground tokens. With the adopted retention ratio $\rho=20\%$, the combined
  design achieves $20.6\times$--$91.1\times$ training speedup and $22.3\times$--$74.6\times$ end-to-end inference speedup within the common foreground-ratio
  range of our dataset.
      }
      \label{fig:efficiency_analysis}

      \vspace{0.8em}

\noindent
\begin{minipage}[c]{0.48\textwidth}
    \centering

    \includegraphics[width=\linewidth]
        {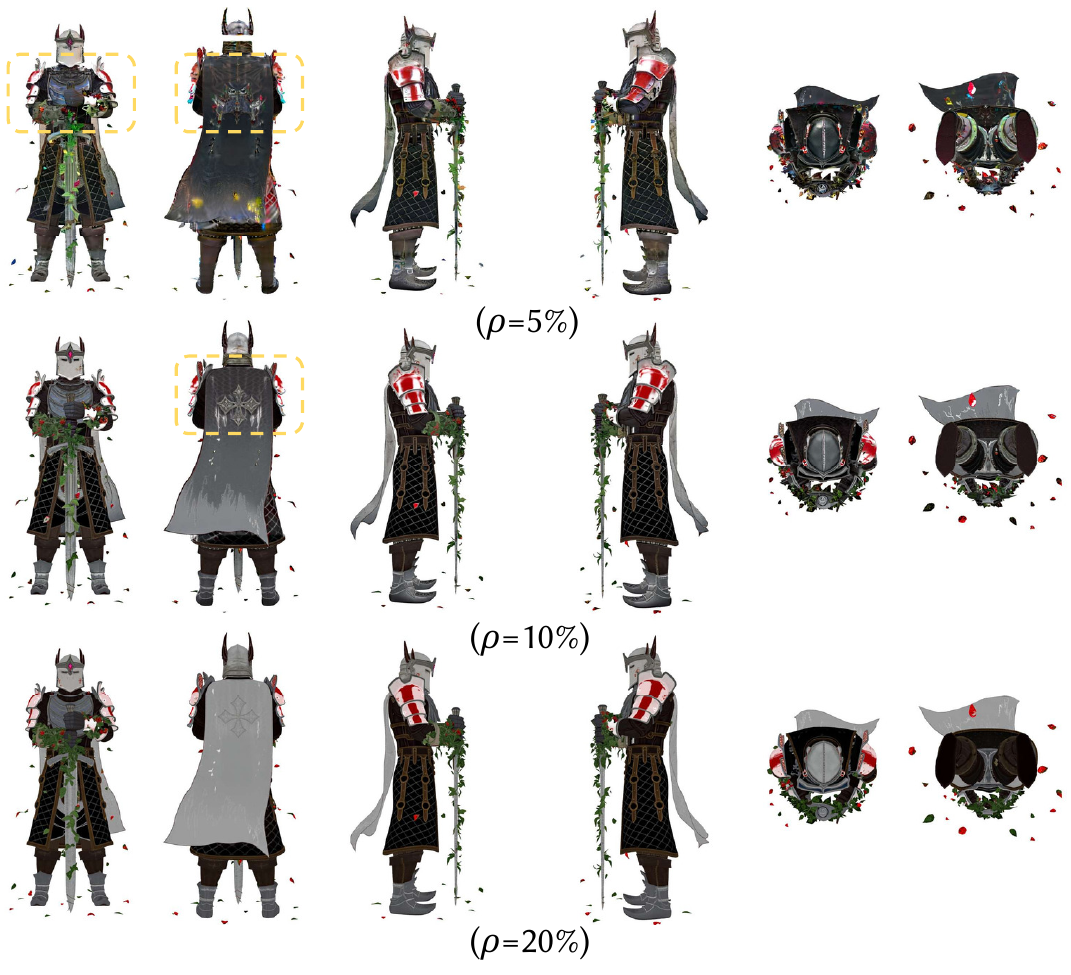}
    \captionof{figure}{
        Effect of the Block-Sparse Attention retention ratio.
        As the retention ratio $\rho$ decreases, fewer key blocks
        are retained during inference, leading to progressively
        lower visual quality.
    }
    \label{fig:top-k-select}

    \vspace{0.8em}

    \includegraphics[width=\linewidth]
        {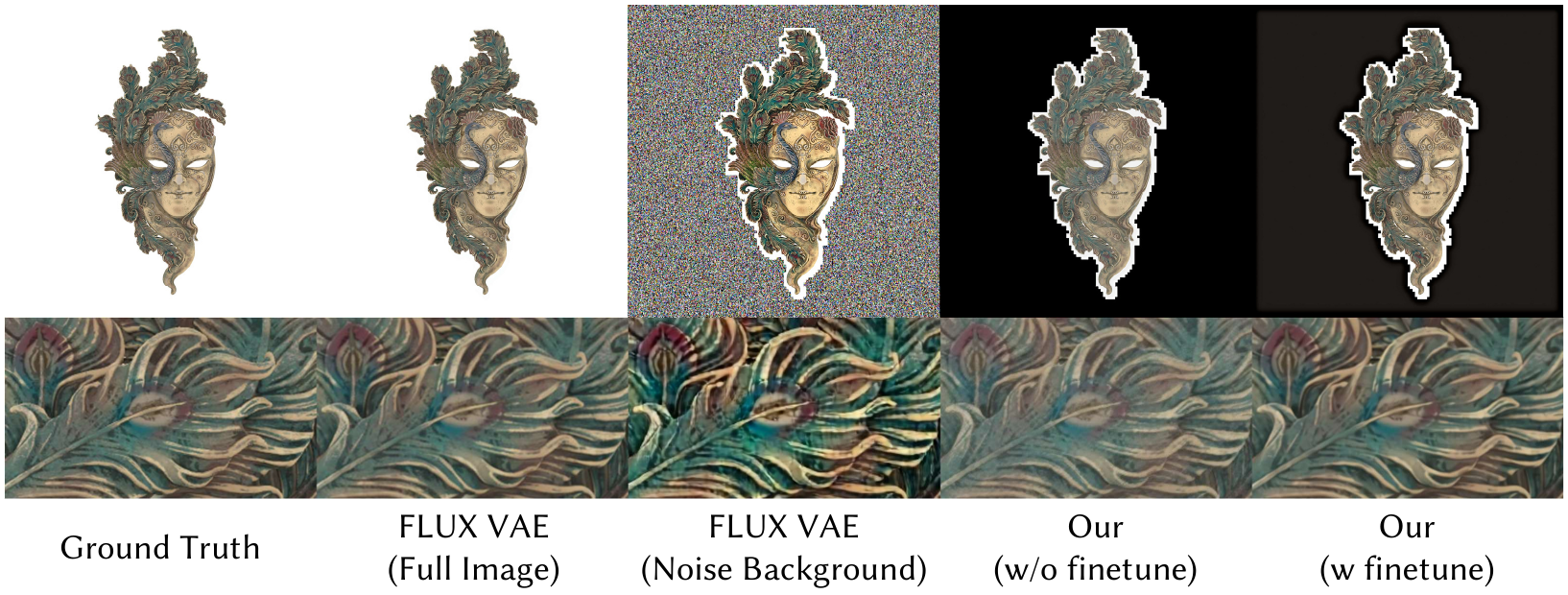}
    \captionof{figure}{
        Effect of Foreground-Aware VAE Decoding.
        Directly decoding a composite latent with denoised foreground
        and noisy background causes noticeable foreground quality
        degradation. Replacing the noisy background with a canonical
        VAE background latent, together with foreground-restricted
        decoder fine-tuning, improves reconstruction fidelity.
    }
    \label{fig:vae_recon}
\end{minipage}
\hfill
\begin{minipage}[c]{0.48\textwidth}
    \centering

    \includegraphics[width=0.95\linewidth]
        {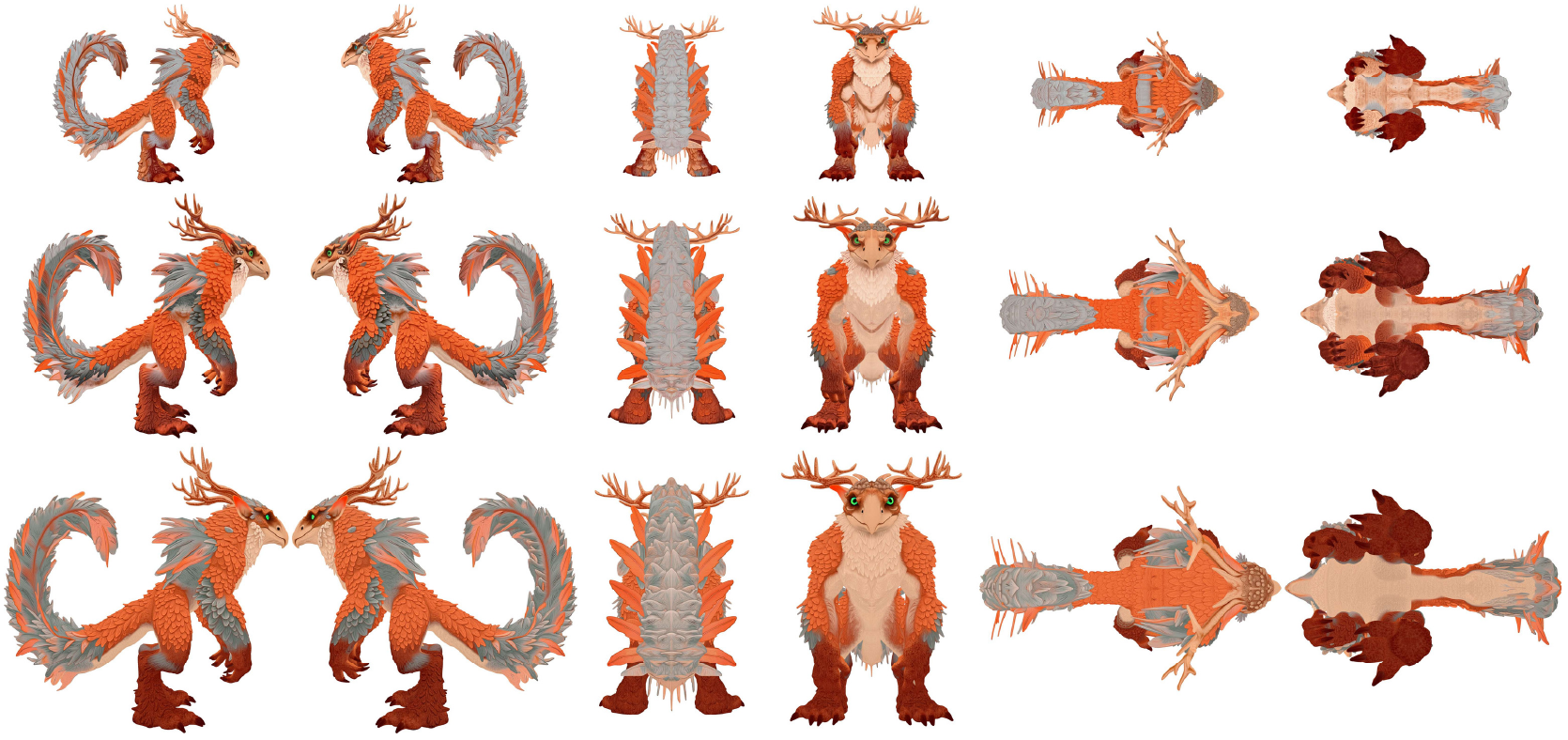}
    \captionof{figure}{
        Test-time foreground scaling.
        By increasing the foreground ratio of the geometry condition
        at inference time, UltraTex retains more foreground tokens
        during denoising. This increases the valid foreground area
        in the generated multi-view observations, yielding more
        texture evidence for subsequent 3D texturing.
    }
    \label{fig:test_time_foreground_scaling}

    \vspace{0.8em}

    \includegraphics[width=0.9\linewidth]
        {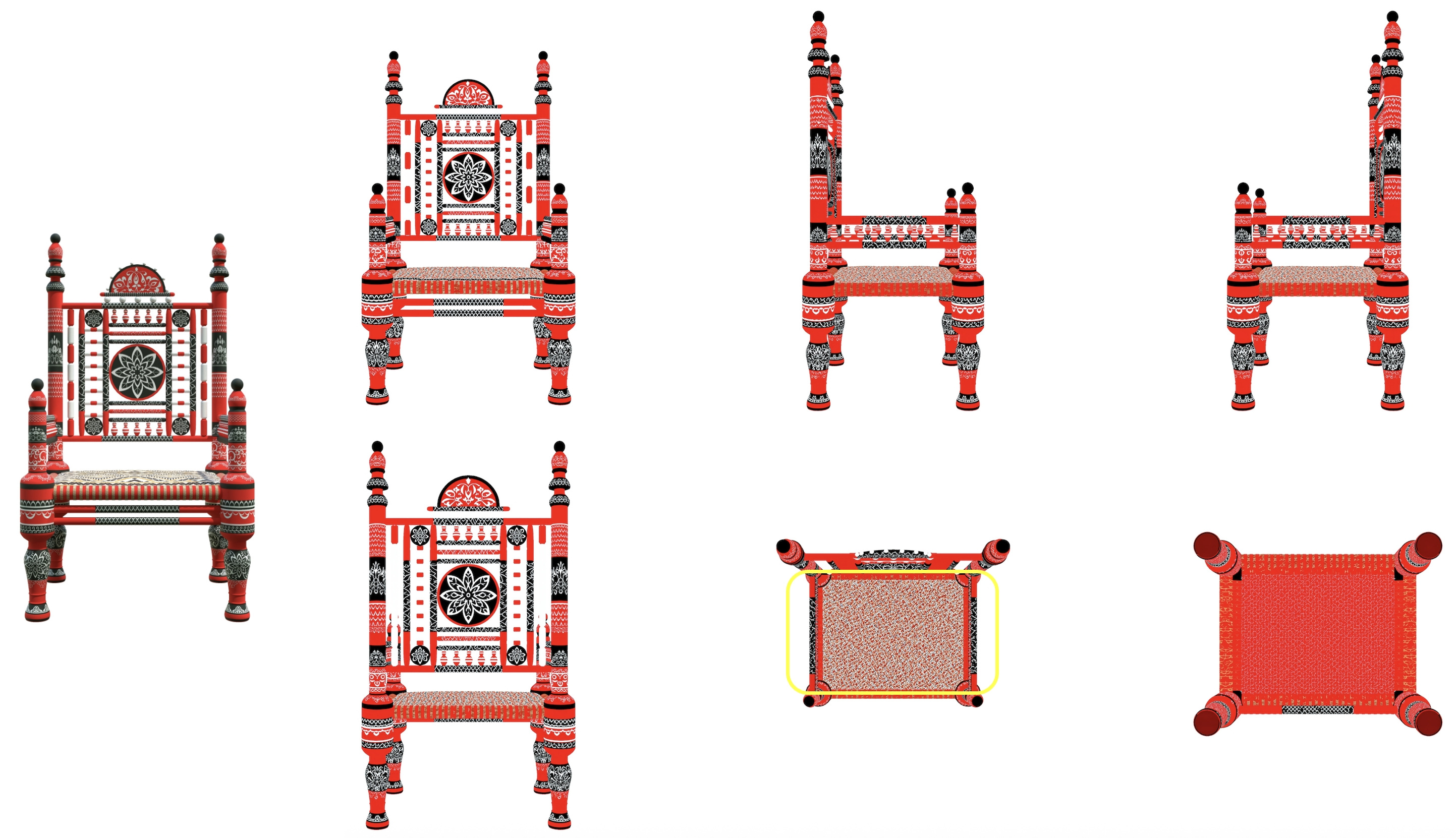}
    \captionof{figure}{
        Failure case. Our method may struggle with objects containing
        highly repetitive texture patterns.
    }
    \label{fig:fail_case}
\end{minipage}

      \vspace{0.8em}

      \begin{minipage}[t]{0.48\textwidth}

      \end{minipage}\hfill
      \begin{minipage}[t]{0.48\textwidth}

      \end{minipage}
  \end{figure*}

We evaluate the efficiency of UltraTex's two core components---Background Token Dropping (BTD) and Block-Sparse Attention (BSA)---at 2048 resolution on a single H200 GPU. 
As shown in \Cref{fig:efficiency_analysis}, BTD effectively reduces the sequence length by dropping background tokens, thereby lowering training cost, inference cost, and peak GPU memory compared with full-sequence processing. 
Building on the compressed foreground sequence, BSA further reduces attention computation among retained tokens, with smaller retention ratios $\rho$ yielding greater speedups. Within the dataset's typical foreground ratio range (10\%--30\%), our adopt $\rho=20\%$ achieves \textbf{20.6$\times$--91.1$\times$} training speedup and \textbf{22.3$\times$--74.6$\times$} end-to-end inference speedup. Overall, the combination of BTD and BSA effectively alleviates the efficiency bottleneck of high-resolution multi-view diffusion.

\paragraph{Selection of Block-Sparse Attention Retention Ratio.}
The retention ratio $\rho$ in BSA controls the trade-off between computational efficiency and generation quality.
A smaller $\rho$ yields higher speedup but may discard important token interactions, leading to texture loss and visual degradation.
We analyze the effect of $\rho$ through validation experiments on a low-resolution full-attention model trained with BTD.
Specifically, we replace full attention with BSA using different retention ratios at inference time.
As shown in \Cref{fig:top-k-select}, as the retention ratio $  \rho  $ decreases, the generation quality also declines; among which $  \rho=5\%  $ leads to noticeable texture degradation.
Although smaller $  \rho  $ provides greater acceleration on long sequences, the majority of our training samples have a foreground ratio below 30\%.
Under this distribution, BTD already substantially shortens the effective sequence length.
Thus, an overly aggressive $\rho$ brings only limited additional efficiency gains while increasing the risk of quality degradation.
Based on this quality--efficiency trade-off, we set $\rho=20\%$ in our final model.

\subsection{Test-Time Foreground Scaling}
UltraTex’s foreground-only denoising naturally supports inference-time control over the fraction of generated pixels allocated to the object foreground. We call this controllable adjustment \textit{Test-Time Foreground Scaling}. As shown in \Cref{fig:test_time_foreground_scaling}, by increasing the Background Token Dropping retains more foreground tokens during generation, thereby increasing the valid foreground area in the generated multi-view observations and yielding more texture evidence for subsequent 3D texturing.


\section{Conclusion}
We presented UltraTex, an efficient end-to-end multi-view diffusion framework that generates 2048-resolution multi-view texture outputs for high-quality 3D texturing. We observed that high-resolution object-centric multi-view diffusion is limited by both background-induced sequence redundancy and attention redundancy among foreground tokens. Based on this observation, we introduced Background Token Dropping to remove redundant background tokens, and adopted Block-Sparse Attention to reduce dense attention computation over the compressed foreground sequence. To preserve the computational efficiency of foreground-only denoising during inference, we further designed Foreground-Aware VAE Decoding, which ensures high-quality reconstruction of the final high-resolution views from composite foreground-background latents. We also constructed G-buffer TexVerse, a large-scale high-resolution multi-view rendering dataset covering over 268K 3D assets. Extensive experiments demonstrate that UltraTex produces detailed and visually faithful high-resolution multi-view texture outputs, while substantially improving computational efficiency and reducing memory cost, providing an effective solution for high-quality 3D texturing.

\section{Limitations}
Our method still has several limitations. First, it may struggle with objects containing highly repetitive texture patterns. A representative failure case is shown in \Cref{fig:fail_case}.
Second, our framework is inherently constrained by the capability of the pretrained FLUX model on which it is built. This limitation mainly manifests in two aspects. (i) The fidelity of the final multi-view outputs is bounded by the VAE latent representation used by FLUX. Although we fine-tune the VAE decoder for foreground-aware reconstruction, the overall encoder--decoder pipeline still operates through a compressed latent space. Consequently, for high-resolution rendered ground-truth images, the encode--decode process may blur part of the high-frequency details present in the rendered images, thereby limiting the finest texture details that can ultimately be reconstructed. (ii) The effective operating range of FLUX is better aligned with resolutions of approximately 1K--2K, while 4K inputs substantially exceed the resolution range to which the pretrained model is primarily adapted. Therefore, further scaling end-to-end multi-view texture generation to 4K or even 8K resolutions may require redesigning DiT--VAE foundation models that natively support ultra-high-resolution inputs.

\section{Acknowledgements}
This work was supported in part by National Natural Science Foundation of China (No. 62572212, No. 62376060), Science and Technology Development Plan of Jilin Province (No. 20260203049SF), the Fundamental Research Funds for the Central Universities
and Ningbo grant (2025Z038).

\clearpage

\clearpage
\bibliographystyle{ACM-Reference-Format}
\bibliography{main}

\clearpage
\appendix
\nobalance
\setcounter{figure}{0}
\setcounter{table}{0}
\setcounter{equation}{0}
\renewcommand{\thefigure}{S\arabic{figure}}
\renewcommand{\thetable}{S\arabic{table}}
\renewcommand{\theequation}{S\arabic{equation}}
\renewcommand{\theHfigure}{supp.\arabic{figure}}
\renewcommand{\theHtable}{supp.\arabic{table}}
\renewcommand{\theHequation}{supp.\arabic{equation}}

  \makeatletter
  \twocolumn[{%
    \hsize=\textwidth
    \@ACM@title@width=\hsize
    \vbox{\noindent\@titlefont
      \parbox[t]{\@ACM@title@width}{\raggedright
        \@titlefont\noindent
        Supplementary Material 
      }%
      \par\bigskip
    }%
  }]
  \makeatother

\section{Dataset Details}
\label{sec:supp_dataset}

We provide a detailed description of our G-buffer TexVerse dataset, covering its asset curation, rendering protocol, data organization, and foreground-ratio statistics.

\paragraph{3D Asset Curation}
Compared to the commonly used Objaverse dataset, whose maximum texture resolution is limited to 1024, recent datasets such as TexVerse provide large-scale 3D assets with high-resolution textures. 
However, the raw TexVerse still contains a substantial portion of assets unsuitable for training, including geometrically broken 3D scans, models with missing textures, low-quality textures, and other visual or structural defects.

\begin{figure}[t]
    \centering
    \includegraphics[width=\linewidth]{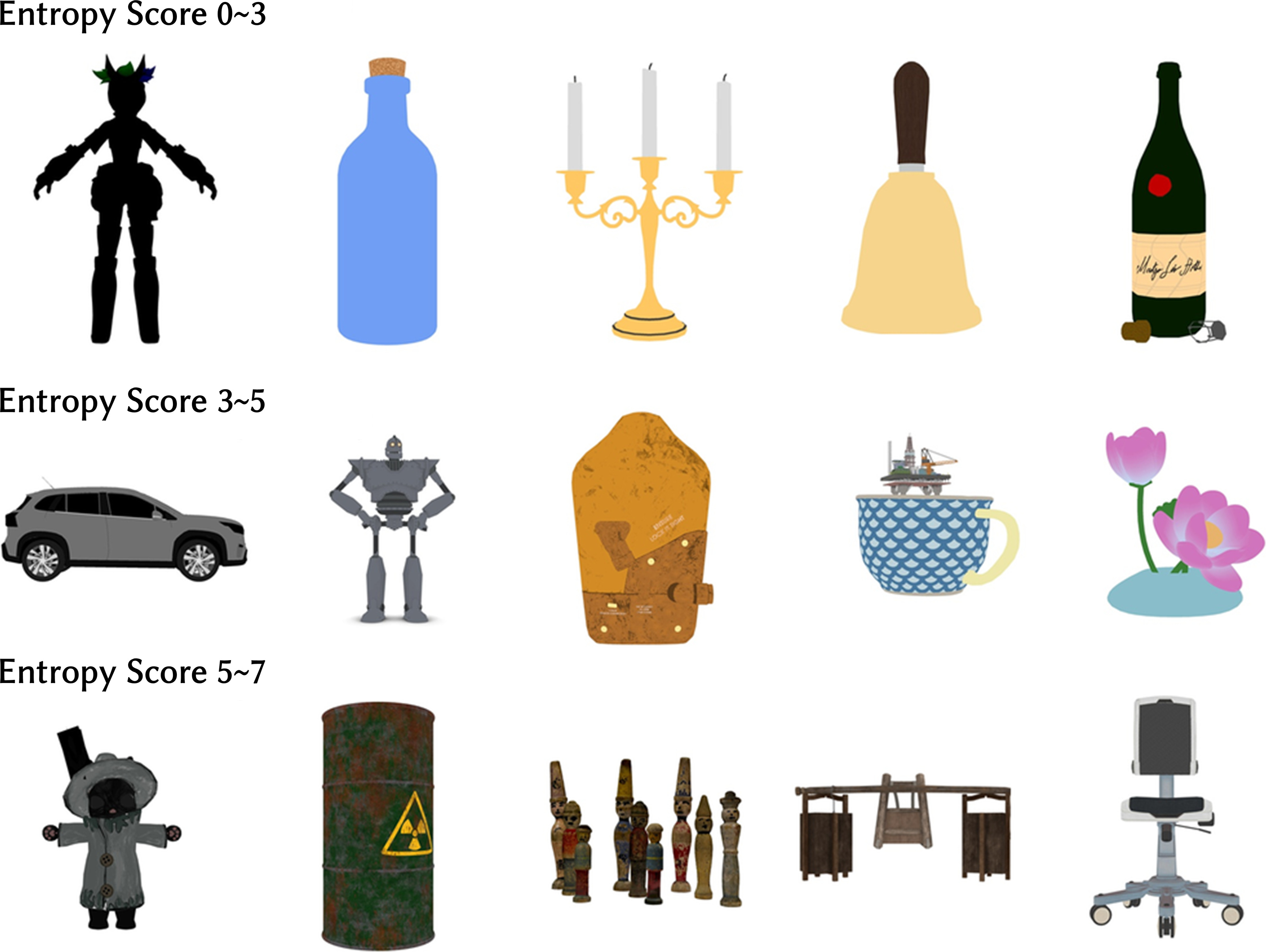}
    \caption{Examples of albedo entropy filtering.}
    \label{fig:albedo_entropy_filtering}
\end{figure}

To obtain reliable training assets, we designed a rigorous multi-stage filtering pipeline. The process is as follows:
\begin{itemize}[itemsep=0.1em, leftmargin=1.5em]
    \item \textbf{Stage 1: Visual Quality Assessment (858K → 402K).} All assets are rendered as low-resolution four-view images and evaluated visual quality using GPT-5. Assets that are unrecognizable, broken, incomplete, excessively stretched, improperly scaled, lacking texture or material, or visually simplistic are removed. 
    
    \item \textbf{Stage 2: Non-BSDF Material Filtering  (402K → 348K).} Assets whose materials are not organized under standard PBR/BSDF shader workflows are excluded. This includes models relying on non-standard techniques such as emission-based shaders, which fail to provide valid albedo, roughness, or metallic information and produce uninformative outputs for physically-based rendering pipelines.
    
    \item \textbf{Stage 3: Albedo Entropy Filtering (348K → 297K).} We compute the Shannon entropy of foreground pixels (determined by the alpha channel) in rendered albedo maps across six views. The maximum entropy value among the views served as the texture complexity score for each asset. Assets with entropy in buckets 0--3 (indicating overly simple or uniform albedo appearance) are discarded, while those in buckets 4--7 are retained. We provide representative examples of each bucket in \Cref{fig:albedo_entropy_filtering}.
    
    \item \textbf{Stage 4: AI-Generated Content Filtering (297K → 268K).} Assets tagged as AI-generated in the metadata (including \texttt{meshy}, \texttt{tripo}, and \texttt{createdwithai}) are removed.
\end{itemize}
After applying the complete multi-stage filtering pipeline, over 268K high-quality 3D assets are retained for the final dataset.

\begin{figure}[t]
    \centering
    \includegraphics[width=\linewidth]{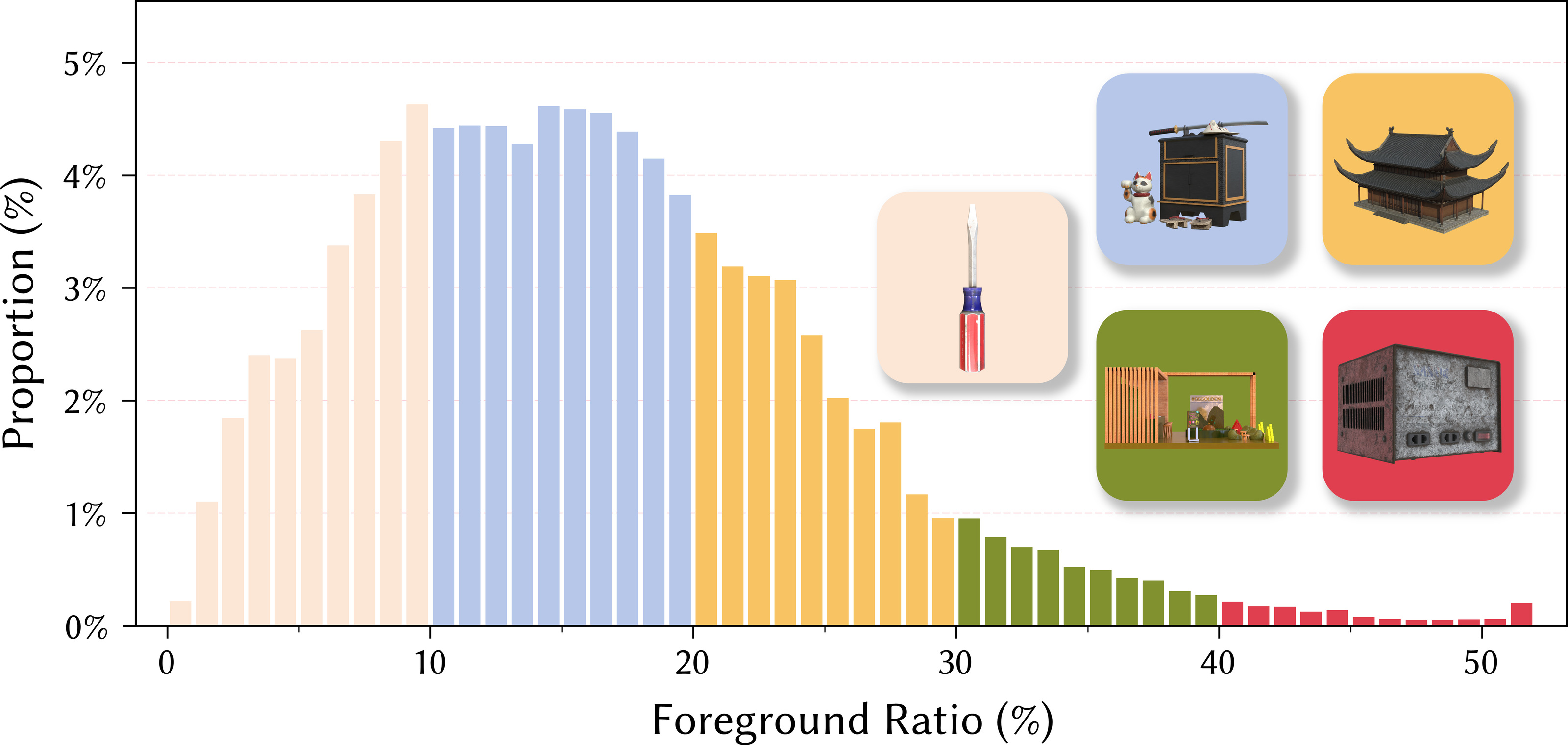}
    \caption{Foreground ratio statistics of our rendered dataset.
    Most objects have relatively small foreground regions, with the distribution concentrated in the $5\%$--$30\%$ range and only a few samples exceeding $40\%$.}
    \label{fig:foreground_ratio_dataset}
\end{figure}

\paragraph{Rendering Protocol and Data Organization.}
We use Blender Cycles to render high-resolution multi-view data for each 3D asset. All rendered outputs, including geometry/material maps and shaded images, preserve an alpha channel to indicate the visible foreground region. 
For each object, we render geometry and material attributes, including shading normal maps, canonical coordinate maps (CCM), albedo maps, and metallic/roughness maps when available. To obtain shaded observations, we construct an illumination pool of 862 HDR environment maps from Poly Haven. Three HDR maps are randomly sampled per object, and shaded images are rendered under each lighting condition using the same camera poses as the attribute renderings.

\paragraph{Camera Configurations.}
Every asset is rendered under two camera configurations, denoted \emph{canonical} and \emph{sphere}. Both cover the same asset set and share an identical rendering setup, including the camera intrinsics, the object normalization, the aspect-ratio-adaptive camera distance, and the three sampled HDR maps; they differ only in the set of viewpoints:
\begin{itemize}[itemsep=0.1em, leftmargin=1.5em]
    \item \textbf{Canonical (6 views).} Six axis-aligned viewpoints, comprising four side views at azimuths ${0^\circ}$, ${90^\circ}$, ${180^\circ}$, and ${270^\circ}$ with zero elevation, plus a top and a bottom view at elevations $\pm 90^\circ$. \textbf{This is the configuration used to train and evaluate UltraTex}: it defines the six-view multi-view layout of our diffusion model, and every result, token count, and efficiency measurement reported in the main paper is obtained under it. Unless stated otherwise, all mentions of multi-view data in this work refer to this configuration.
    \item \textbf{Sphere (36 views).} An elevation--azimuth grid of twelve azimuths uniformly spaced every $30^\circ$ at each of three elevations, ${-40^\circ}$, ${-20^\circ}$, and ${30^\circ}$. The azimuth grid subsumes the four canonical side-view azimuths, so the two configurations remain registered and can be used jointly on the same asset. This configuration is \textbf{not used by UltraTex}; we render and release it to support the broader community, as its dense viewpoint coverage of high-resolution G-buffer attributes benefits tasks beyond ours, such as novel-view synthesis, sparse-view and dense-view reconstruction, PBR material estimation, and texture baking or reprojection pipelines that require observations well outside a six-view layout.
\end{itemize}
For the canonical configuration, we additionally render training reference images under the same three HDR lighting conditions.
For each lighting condition, we sample four random front-facing reference views from four angular regions, with azimuth \(\alpha\) and elevation \(\beta\) sampled as
\(
(\alpha,\beta) \in
\left([-45^\circ,-8^\circ]\cup[8^\circ,45^\circ]\right)
\times
\left([-5^\circ,0^\circ]\cup[5^\circ,20^\circ]\right).
\)
The camera distance is adaptively set according to the object aspect ratio to ensure full visibility, with an additional perturbation \(\delta \in [-0.1,0.1]\) applied to reference views for diversity.
The rendering resolution is determined by the original texture resolution of each asset: assets with \(1024\)-resolution textures are rendered at \(2048 \times 2048\), while those with texture resolutions greater than or equal to \(2048\) are rendered at \(4096 \times 4096\). These two groups account for \(26.39\%\) and \(73.61\%\) of the dataset, respectively. 
During our training, all rendered views are resized to \(2048 \times 2048\), and only the normal maps are used as geometry conditions.

\paragraph{Foreground Ratio Statistics}
The foreground ratio distribution across our rendered dataset is sharply concentrated and right-skewed, as shown in \Cref{fig:foreground_ratio_dataset}. For each object, this ratio is defined as the average proportion of non-background pixels across the six albedo views. Across all 268K models, more than $85\%$ of the models fall within the $5\%$--$30\%$ range, fewer than $1.2\%$ exceed $40\%$, and the maximum observed ratio is $51.2\%$. Consequently, in nearly all samples, the textured foreground occupies only a small and tightly clustered portion of the canvas, indicating substantial redundancy in object-centric renderings.

\section{Detailed Efficiency Analysis}
\label{sec:supp_efficiency}

\Cref{tab:foreground_ratio_efficiency} reports the per-foreground-ratio efficiency measurements at $2048$ resolution on a single NVIDIA H200 GPU. We sweep the foreground ratio from $10\%$ to $100\%$ and record absolute training step time and end-to-end inference time under dense FlashAttention and Block-Sparse Attention (BSA) with $\rho\in\{20\%,10\%,5\%\}$ on top of the sequence shortened by Background Token Dropping (BTD) (\Cref{tab:training_step_time,tab:inference_e2e_time}).

The key addition is the \emph{per-component} speedup decomposition in \Cref{tab:training_speedup,tab:inference_speedup}. The total acceleration factors as
\begin{equation*}
    \mathrm{Joint} \;=\; \underbrace{\mathrm{BTD\text{-}only}}_{\text{sequence shortening}} \;\times\; \underbrace{\mathrm{Sparse}\ \mathrm{after}\ \mathrm{BTD}}_{\text{attention sparsification}},
\end{equation*}
and the two columns isolate each mechanism's contribution across the full foreground sweep. BTD alone delivers $1.4\times$--$53.0\times$ training speedup ($1.0\times$--$48.1\times$ inference) and dominates at low foreground ratios where background tokens are the bulk of the sequence. BSA after BTD contributes a steadier $1.7\times$--$2.9\times$ training ($1.6\times$--$4.1\times$ inference) and grows with the foreground ratio, complementing BTD when there is less background to drop. The Joint column reproduces the $20.6\times$--$91.1\times$ training and $22.3\times$--$74.6\times$ inference ranges quoted in the main paper, broken down by foreground bin.

\begingroup
\setlength{\intextsep}{5pt}
\captionsetup{skip=3pt}
\section{Influence of the Pretrained Foundation Model}
\label{sec:supp_base_model}

Replacing FLUX with FLUX.2-dev as the pretrained foundation model markedly improves reference identity preservation in our qualitative comparison.
As shown in \Cref{fig:base_model_comparison}, the FLUX-based result captures the overall structure of the bulletin board but alters distinctive details in its posters, illustrations, and central table.
The FLUX.2-dev-based result more faithfully reproduces these visual elements and their spatial arrangement, retaining a closer resemblance to the reference image.
This example suggests that upgrading the underlying generative model can further improve the fidelity of image-guided 3D texturing, particularly for assets whose identity depends on fine-grained texture content.

\begin{figure}[H]
    \centering
    \includegraphics[width=\linewidth]{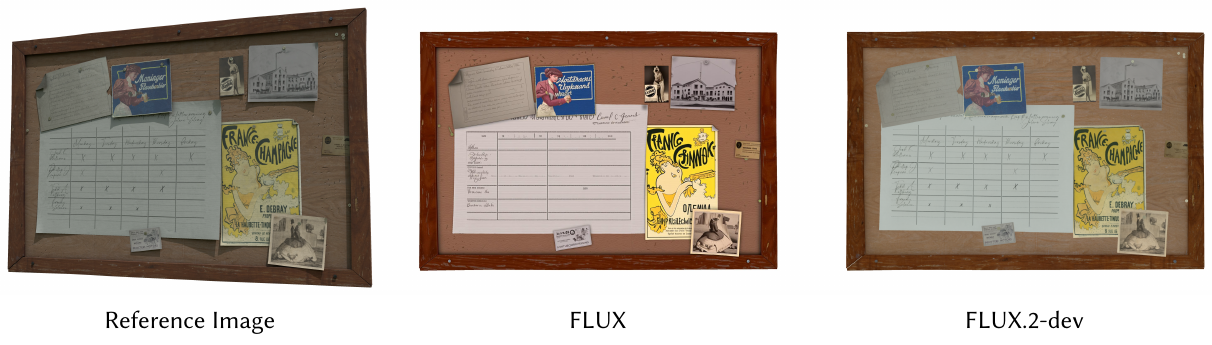}
    \caption{Influence of the pretrained foundation model on identity preservation. From left to right: the reference image, the result using FLUX, and the result using FLUX.2-dev. FLUX.2-dev more faithfully preserves the distinctive posters, illustrations, and table layout of the reference.}
    \Description{Three images of a bulletin board. The left image is the reference, the middle image is the FLUX result, and the right image is the FLUX.2-dev result. The right image retains the reference's poster designs and central table more faithfully than the middle image.}
    \label{fig:base_model_comparison}
\end{figure}

\section{Extension to PBR Material Generation}
\label{sec:supp_pbr}

Our pipeline naturally extends to physically based rendering (PBR) material generation.
With FLUX.2-dev as the pretrained foundation model, we adapt the image-guided multi-view formulation to jointly predict roughness and metallic maps conditioned on a reference image and the object's geometry.
\Cref{fig:pbr_extension} shows the reference image alongside six generated views of the packed roughness--metallic maps.

\begin{figure}[H]
    \centering
    \includegraphics[width=\linewidth]{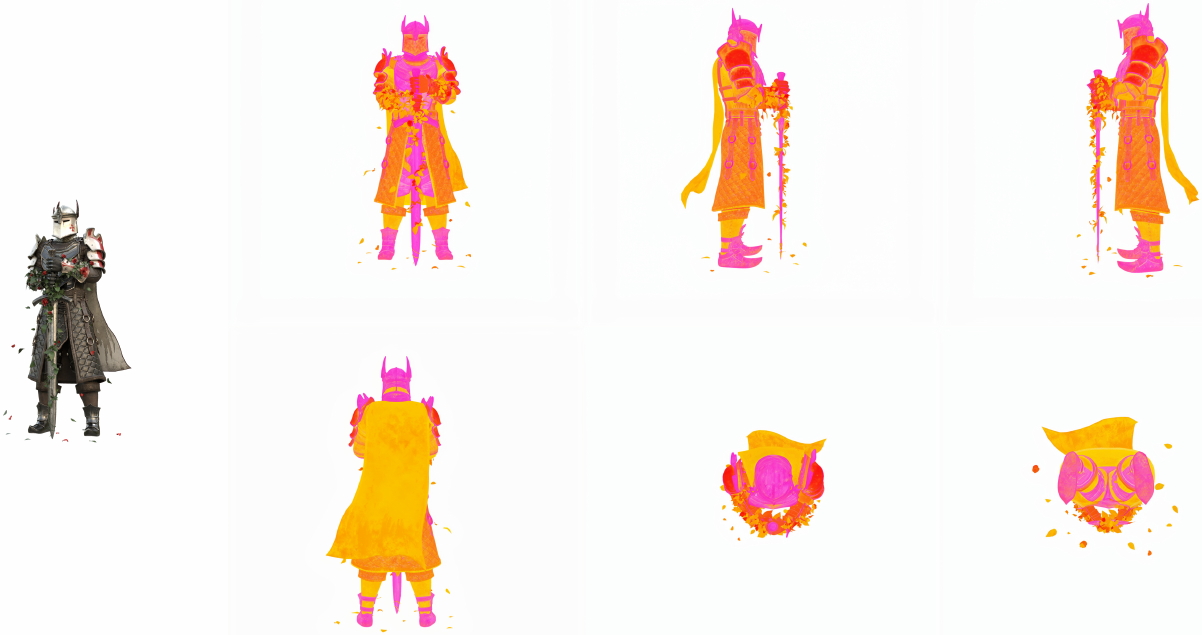}
    \caption{PBR material generation with FLUX.2-dev as the pretrained foundation model. The reference image (left) conditions the generation of six-view roughness--metallic maps (right), visualized as packed material channels.}
    \Description{A reference image of an armored character on the left, vertically centered next to a two-row, three-column grid of generated roughness and metallic maps on the right.}
    \label{fig:pbr_extension}
\end{figure}

\endgroup
\onecolumn

\begin{table*}[t]
\centering
\caption{Efficiency analysis of UltraTex at 2048 resolution under different foreground ratios.}
\label{tab:foreground_ratio_efficiency}

\begin{subtable}{\textwidth}
\centering
\caption{Training step time.}
\label{tab:training_step_time}
\resizebox{\textwidth}{!}{
\begin{tabular}{cccccc}
\toprule
Foreground Ratio & Token Length & Flash Attn & Sparse Attn ($\rho$=20\%) & Sparse Attn ($\rho$=10\%) & Sparse Attn ($\rho$=5\%) \\
\midrule
10\%  & 21,810  & 8.03   & 4.67   & 4.14   & 3.88  \\
20\%  & 43,108  & 24.54  & 11.62  & 9.52   & 8.45  \\
30\%  & 64,409  & 49.74  & 20.66  & 16.09  & 13.85 \\
40\%  & 85,707  & 83.43  & 32.67  & 24.38  & 20.26 \\
50\%  & 107,008 & 124.65 & 46.37  & 33.57  & 27.12 \\
60\%  & 128,306 & 177.08 & 62.97  & 44.31  & 35.20 \\
70\%  & 149,604 & 237.03 & 82.06  & 56.20  & 43.49 \\
80\%  & 170,905 & 299.82 & 102.89 & 69.62  & 53.11 \\
90\%  & 192,203 & OOM    & OOM    & OOM    & OOM   \\
100\% & 213,504 & OOM    & OOM    & OOM    & OOM   \\
\bottomrule
\end{tabular}
}
\end{subtable}

\vspace{0.8em}

\begin{subtable}{\textwidth}
\centering
\caption{Inference end-to-end time.}
\label{tab:inference_e2e_time}
\resizebox{\textwidth}{!}{
\begin{tabular}{lccccc}
\toprule
Foreground Ratio & Token Length & Flash Attn & Sparse Attn ($\rho$=20\%) & Sparse Attn ($\rho$=10\%) & Sparse Attn ($\rho$=5\%) \\
\midrule
10\%  & 21,810  & 53.13   & 34.28  & 32.30  & 31.30  \\
20\%  & 43,108  & 146.04  & 71.03  & 63.74  & 59.84  \\
30\%  & 64,409  & 284.43  & 114.80 & 98.19  & 90.03  \\
40\%  & 85,707  & 466.47  & 166.18 & 137.40 & 122.91 \\
50\%  & 107,008 & 706.67  & 226.43 & 180.87 & 158.08 \\
60\%  & 128,306 & 967.55  & 293.47 & 227.60 & 196.28 \\
70\%  & 149,604 & 1282.06 & 364.50 & 275.75 & 231.47 \\
80\%  & 170,905 & 1644.69 & 442.15 & 328.53 & 270.80 \\
90\%  & 192,203 & 2054.43 & 529.06 & 385.03 & 311.96 \\
100\% & 213,504 & 2556.16 & 628.12 & 446.61 & 356.65 \\
\bottomrule
\end{tabular}
}
\end{subtable}

\vspace{0.8em}

\begin{subtable}[t]{0.48\textwidth}
\centering
\caption{Component-wise training speedup with Sparse Attn ($\rho$=20\%).}
\label{tab:training_speedup}
\resizebox{\linewidth}{!}{
\begin{tabular}{cccc}
\toprule
Foreground Ratio & BTD-only & Sparse after BTD & Joint \\
\midrule
10\%  & 52.98$\times$ & 1.72$\times$ & 91.10$\times$ \\
20\%  & 17.33$\times$ & 2.11$\times$ & 36.61$\times$ \\
30\%  & 8.55$\times$  & 2.41$\times$ & 20.59$\times$ \\
40\%  & 5.10$\times$  & 2.55$\times$ & 13.02$\times$ \\
50\%  & 3.41$\times$  & 2.69$\times$ & 9.17$\times$  \\
60\%  & 2.40$\times$  & 2.81$\times$ & 6.76$\times$  \\
70\%  & 1.79$\times$  & 2.89$\times$ & 5.18$\times$  \\
80\%  & 1.42$\times$  & 2.91$\times$ & 4.13$\times$  \\
90\%  & OOM & OOM & OOM \\
100\% & OOM & OOM & OOM \\
\bottomrule
\end{tabular}
}
\end{subtable}
\hfill
\begin{subtable}[t]{0.48\textwidth}
\centering
\caption{Component-wise E2E inference speedup with Sparse Attn ($\rho$=20\%).}
\label{tab:inference_speedup}
\resizebox{\linewidth}{!}{
\begin{tabular}{cccc}
\toprule
Foreground Ratio & BTD-only & Sparse after BTD & Joint \\
\midrule
10\%  & 48.11$\times$ & 1.55$\times$ & 74.57$\times$ \\
20\%  & 17.50$\times$ & 2.06$\times$ & 35.99$\times$ \\
30\%  & 8.99$\times$  & 2.48$\times$ & 22.27$\times$ \\
40\%  & 5.48$\times$  & 2.81$\times$ & 15.38$\times$ \\
50\%  & 3.62$\times$  & 3.12$\times$ & 11.29$\times$ \\
60\%  & 2.64$\times$  & 3.30$\times$ & 8.71$\times$  \\
70\%  & 1.99$\times$  & 3.52$\times$ & 7.01$\times$  \\
80\%  & 1.55$\times$  & 3.72$\times$ & 5.78$\times$  \\
90\%  & 1.24$\times$  & 3.88$\times$ & 4.83$\times$  \\
100\% & 1.00$\times$  & 4.07$\times$ & 4.07$\times$  \\
\bottomrule
\end{tabular}
}
\end{subtable}

\vspace{0.3em}

\end{table*}


\end{document}